%% file: iclr2026_conference.tex
\documentclass{article} 
\usepackage{iclr2026_conference,times}

\input{math_commands.tex}

\usepackage{hyperref}
\usepackage{booktabs}
\usepackage{tikz}
\usepackage{caption}
\usepackage{multirow}
\usepackage{graphicx}
\usepackage{wrapfig}
\usepackage{tcolorbox}

\title{IL3D: A Large-Scale Indoor Layout Dataset for LLM-Driven 3D Scene Generation}

\author{Wenxu Zhou$^{1}$, Kaixuan Nie$^{2}$, Hang Du$^{2}$, Dong Yin$^{1}$, Wei Huang$^{2}$, Siqiang Guo$^{2}$\\
\textbf{Xiaobo Zhang$^{2}$, Pengbo Hu$^{2,\dagger}$}\\
$^1$University of Science and Technology of China\quad $^2$Songying Technology\\
}

\iclrfinalcopy 
\begin{document}

\maketitle
\input{sec/0_abstract}
\input{sec/1_introduction}
\input{sec/2_related_work}
\input{sec/3_the_il3d_dataset}
\input{sec/4_dataset_analysis}
\input{sec/5_experiments}
\input{sec/6_conclusion}

\bibliography{iclr2026_conference}
\bibliographystyle{iclr2026_conference}

\appendix
\input{sec/7_appendix}

\end{document}

%% file: math_commands.tex
\usepackage{amsmath,amsfonts,bm}

\def\eqref#1{equation~\ref{#1}}

\def\1{\bm{1}}

\DeclareMathAlphabet{\mathsfit}{\encodingdefault}{\sfdefault}{m}{sl}
\SetMathAlphabet{\mathsfit}{bold}{\encodingdefault}{\sfdefault}{bx}{n}



%% file: sec/0_abstract.tex
\begin{abstract}
In this study, we present IL3D, a large-scale dataset meticulously designed for large language model (LLM)-driven 3D scene generation, addressing the pressing demand for diverse, high-quality training data in indoor layout design. Comprising 27,816 indoor layouts across 18 prevalent room types and a library of 29,215 high-fidelity 3D object assets, IL3D is enriched with instance-level natural language annotations to support robust multimodal learning for vision-language tasks. We establish rigorous benchmarks to evaluate LLM-driven scene generation. Experimental results show that supervised fine-tuning (SFT) of LLMs on IL3D significantly improves generalization and surpasses the performance of SFT on other datasets. IL3D offers flexible multimodal data export capabilities, including point clouds, 3D bounding boxes, multiview images, depth maps, normal maps, and semantic masks, enabling seamless adaptation to various visual tasks. As a versatile and robust resource, IL3D significantly advances research in 3D scene generation and embodied intelligence, by providing high-fidelity scene data to support environment perception tasks of embodied agents. The dataset and accompanying code are publicly accessible\footnote{The project homepage is located at \url{https://wenxuzhou.github.io/project/IL3D/}}.
\end{abstract}

%% file: sec/1_introduction.tex
\section{Introduction}
3D indoor scene generation has emerged as a pivotal technology bridging embodied intelligence, smart home design, virtual reality interaction, and robotic environmental perception. Its core objective is to transform abstract spatial requirements into physically plausible and semantically coherent indoor layouts~\cite{yang2024holodeck,ccelen2024design,zhou2025roomcraft}. In this domain, large language models (LLMs), leveraging their robust natural language understanding and reasoning capabilities, have become critical tools for driving scene generation. However, the precise modeling of indoor scenes by LLMs heavily relies on high-quality synthetic datasets~\cite{khanna2024habitat,kolve2017ai2}. These datasets must not only encompass diverse indoor scene types but also provide fine-grained information to support 3D perception tasks, including object geometry, semantic relationships, and multimodal representations. Such data is essential to address challenges like object overlap, boundary overflow, and other physical plausibility issues, as well as to meet the requirements of 3D point cloud analysis, semantic segmentation, and related perception tasks.

While existing mainstream indoor scene synthesis datasets~\cite{khanna2024habitat, fu20213d} have advanced 3D perception research to some extent, they exhibit notable limitations. Some datasets focus primarily on accumulating furniture assets but lack sufficient coverage of diverse indoor scene types, making it challenging to support modeling for varied living spaces such as bedrooms, kitchens, or study rooms. Others provide interactive 3D scenes but lack fine-grained annotations, failing to deliver critical information such as object materials, poses, or spatial relationships, which leads to biases in LLMs' semantic understanding of indoor scenes. Additionally, certain datasets prioritize visual realism but fall short in supporting the data formats required for multimodal 3D perception tasks, such as depth maps, normal maps, or semantic masks, thereby limiting their utility for training perception models and hindering the synergy between 3D perception and scene generation tasks.

Furthermore, 3D perception tasks for indoor scenes impose stringent requirements on the “functionality” and “diversity” of datasets. On one hand, datasets must ensure that indoor layouts adhere to real-world functional logic (e.g., appropriate placement of cabinets and appliances in kitchens or beds and wardrobes in bedrooms) to enable perception models to understand the functional context of indoor scenes. On the other hand, datasets need to cover indoor scenes with varying areas and object densities to prevent biases that could degrade the generalization ability of perception models in complex scenarios, such as densely populated living rooms or compact bathrooms. These demands underscore the urgent need for a large-scale synthetic dataset tailored to indoor scenes that addresses the requirements of 3D perception tasks.

To address these challenges, this study introduces the IL3D dataset, a large-scale indoor layout dataset designed to meet the core needs of indoor scene generation and 3D perception. By integrating high-quality existing scene resources and supplementing them with targeted synthetic data, IL3D overcomes the shortcomings of current datasets in terms of indoor scene diversity, annotation completeness for 3D perception, and adaptability to multiple tasks. The dataset provides training samples closely aligned with real-world indoor environments for LLMs while offering multimodal data, including semantic point clouds, 3D bounding boxes, and multi-view RGB images, to support 3D perception tasks. This facilitates a synergistic improvement in both the quality of indoor scene generation and the accuracy of 3D perception, laying a robust data foundation for subsequent academic research and industrial applications related to indoor scenes.
Our contributions can be summarized as follows:
\begin{itemize}
\item We introduce the IL3D dataset, containing 27,816 indoor layouts and 29,215 indoor object models, effectively meeting the data needs of diverse 3D scene understanding tasks.
\item The dataset provides instance-level natural language descriptions and supports multiple data formats, including semantic point clouds, 3D bounding boxes, multi-view RGB images, depth maps, normal maps, and semantic masks, ensuring compatibility with a wide range of downstream visual tasks.
\item Experiments demonstrate that direct supervised fine-tuning (SFT) on IL3D significantly improves the performance of LLM-driven layout generation, highlighting its critical role in generative tasks for indoor 3D scene synthesis.
\item Constructed primarily from USDZ-format assets and USDA-format scenes, the dataset is readily parsed and analyzed by LLMs, ensuring seamless integration with mainstream graphics processing and simulation software.
\end{itemize}

%% file: sec/2_related_work.tex
\section{Related Work}
The development of large-scale, high-fidelity indoor scene datasets is critical for advancing 3D scene understanding, embodied intelligence, and 3D scene generation, as these datasets provide diverse, semantically rich room and layout representations to support tasks such as object navigation, scene layout generation, and interactive simulation.
\subsection{Synthesis Scene Dataset}
3D-FRONT~\cite{fu20213d} focuses on furniture resources, providing an asset library with 13,151 high-quality 3D object models and 6,813 synthetic houses. Its modular object component design enables flexible scene assembly, laying a solid foundation for the subsequent construction of full-scene datasets. Building on 3D-FRONT, 3D-FUTURE~\cite{fu20213d_} further enhances data adaptability by offering 9,938 high-quality 3D CAD furniture models with texture and semantic annotations. It supports the export of color images, semantic masks, depth maps, and normal maps, significantly expanding the dataset’s application in visual tasks like scene synthesis and texture transfer. MetaScenes~\cite{yu2025metascenes} provides 10,245 high-fidelity indoor scenes with multimodal annotations (semantic, geometric, and natural language descriptions) and dynamic interaction support. It optimizes scene realism and interactivity, making it particularly suitable for multimodal learning and virtual reality applications, and opens up new possibilities for 3D scene generation and embodied intelligence research.

CHOrD~\cite{su2025chord} achieves breakthroughs in scale and controllability, including 9,706 collision-free house-scale indoor scenes with hierarchical layouts and customizable floor plans. Through advanced optimization techniques for fine-tuning object placement and spatial organization, it enables fine-grained control over scene topology, serving as a robust foundation for digital twin generation and large-scale simulation research.

AI2-THOR~\cite{kolve2017ai2} integrates sub-datasets including iTHOR, RoboTHOR~\cite{deitke2020robothor}, ProcTHOR~\cite{deitke2022}, and ArchitecTHOR~\cite{eftekhar2023selective}, covering 120 large-scale interactive 3D indoor scenes (e.g., kitchens, bedrooms, bathrooms, and living rooms). With over 2,000 unique interactive objects rendered via the Unity engine, it supports visual question answering (VQA)~\cite{antol2015vqa} and physics-based simulation experiments, providing critical support for embodied intelligence research. HSSD~\cite{khanna2024habitat} offers 211 photorealistic 3D scenes based on real-world floor plans, containing 18,656 3D object assets. By balancing scene diversity and physical plausibility, it enhances the generalization of agents from simulated to real-world environments in object goal navigation tasks, outperforming many previous synthetic datasets.

\subsection{Text in 3D Scene}
Scan2Cap~\cite{chen2021scan2cap} proposes an RGB-D scan-based context-aware dense annotation task for 3D point cloud scenes, aiming to predict bounding boxes and their natural language descriptions for objects. Using an end-to-end training framework, it leverages an attention mechanism to generate descriptive text and a message-passing graph module to capture inter-object spatial relationship features. On the ScanRefer dataset~\cite{chen2020scanrefer}, it achieves a 27.61\% improvement in CiDEr@0.5IoU score compared to 2D baseline methods, laying an important foundation for dense annotation of 3D scenes. ExCap3D~\cite{yeshwanth2025excap3d} proposes an expressive 3D annotation task to generate multi-level 3D object descriptions, including high-level object descriptions and low-level part attribute descriptions. Based on the ScanNet++ dataset~\cite{yeshwanth2023scannet++}, it uses vision-language models to generate 190,000 multi-view annotations (covering 34,000 objects and 947 indoor scenes). By ensuring semantic consistency of constrained text and textual similarity in latent space, it significantly improves description quality, with CiDEr scores outperforming existing methods.

ScanQA~\cite{azuma2022scanqa} focuses on visual question answering (VQA) in 3D scenes, constructing open-ended question-answer pairs based on the ScanNet dataset~\cite{dai2017scannet}. These pairs cover tasks such as object recognition, spatial relationship understanding, and scene context comprehension. By fusing point cloud features with language prompts, ScanQA enables fine-grained understanding of complex 3D scenes, significantly improving the accuracy of VQA tasks. SQA3D~\cite{ma2022sqa3d} further expands the complexity of 3D scene QA by introducing spatial reasoning and multi-object interaction tasks. Based on RGB-D scans, it provides rich semantic and spatial relationship annotations, supporting diverse tasks ranging from single-object recognition to complex scene reasoning. Through explicit textual relationship prompts, it optimizes multi-object reasoning performance, offering new insights for language-guided 3D scene understanding.
\subsection{LLM-Driven Scene Generation}
Research on text-driven scene generation converts complex scene layouts into structured textual descriptions and leverages the reasoning capabilities of LLM to achieve efficient scene generation.

LayoutGPT~\cite{feng2023layoutgpt} proposes a prompt-based generation framework that effectively processes complex language prompts containing numerical values and spatial relationships. This framework significantly improves the accuracy and semantic consistency of generated layouts. I-Design~\cite{ccelen2024design} adopts a multi-agent system and scene graph generation mechanism, supporting users to generate and visually design targets through natural language interaction. It enables an iterative generation process, transforming user preferences into complete 3D layouts. HOLODECK~\cite{yang2024holodeck} focuses on language-guided 3D embodied AI environment generation. It can create diverse indoor scenes, capture the semantics of complex queries, and support zero-shot object navigation tasks. LayoutVLM~\cite{sun2025layoutvlm} integrates vision-language models with differentiable optimization, proposing an innovative scene layout representation method. It generates physically plausible and semantically consistent layouts from unlabeled 3D assets and language instructions, ensuring the robustness of scenes in terms of physical stability and functionality.

LLplace~\cite{yang2024llplace} optimizes the spatial coherence and functional relationship prediction of LLMs through supervised fine-tuning. It performs exceptionally well in generating layouts for complex indoor environments, excelling in both practicality and semantic accuracy. OptiScene~\cite{yang2025llm} enhances the physical plausibility and visual consistency of generated results through two-stage fine-tuning (supervised fine-tuning followed by direct preference optimization). It performs particularly well in handling diverse room types and complex spatial constraints.

%% file: sec/3_the_il3d_dataset.tex
\section{The IL3D Dataset}
\begin{table}[t]
    \centering
    \caption{Statistics of synthetic datasets for indoor scenes.}
    \label{tab:dataset_characteristics}
    \begin{tabular}{lcccc}
        \toprule
        Dataset & Reference & Number of Rooms & Number of Assets & Text Annotation\\
        \midrule
        3D-FRONT   &   CVPR 2021     & 21.3K                   & 16.6K                    & $\times$ \\
        HSSD      &  CVPR 2024        & 1.1K                    & 18.6K                    & $\times$ \\
        IL3D (Ours)       & -       & 27.8K                   & 29.2K                    & $\checkmark$ \\
        \bottomrule
    \end{tabular}
\end{table}
\subsection{Data}
The IL3D dataset integrates the 3D-FRONT and HSSD datasets, with manual cleaning performed to remove data with abnormal sizes or layouts. We adopted the HOLODECK method to synthesize missing scene types of scene that are missing in these source datasets. The IL3D dataset can be divided into two parts overall: 3D assets and indoor scene layouts. Among them, 3D assets include corresponding object models of different types and multiple types of asset annotations, while scene layouts include information on the position, rotation, and scaling of each object in the scene, as well as the corresponding range of the room.

We constructed the dataset based on the Universal Scene Description (USD) format: using 3D object assets in USDZ format and room layouts in USDA format. The most prominent feature of this format is text readability, meaning large language models can directly read information about objects in the scene from 3D scene models; relevant details are provided in Appendix \ref{3d_retrieval}. Considering that current synthetic scene datasets generally lack natural language annotations, in the IL3D dataset, we annotated each 3D object asset with multiple annotations at different levels. Specifically, coarse labels and fine labels refer to the category names of objects at different hierarchical levels. Additionally, we used Qwen3-VL to annotate scenes with more detailed instance-level descriptions based on multi-view images, including the specific type of the object, appearance description, approximate weight, original pose, constituent materials, and spatial relationships of the object’s layout.
\subsection{Metrics}
The evaluation metrics of the IL3D dataset are divided into two categories: objective metrics and subjective metrics, which are used to comprehensively measure the quality and practicality of generated scenes. Objective metrics include Out-of-Bound (OOB), which is used to detect whether objects in the scene exceed the room boundaries; Object Overlap Rate (OOR), which is used to evaluate the overlap degree between objects; Generation Success Rate (GSR), which is used to calculate the success rate of scene generation; and CLIP-Similarity (CLIP-Sim), which measures the semantic similarity between the generated scene and the reference scene based on the CLIP model.

Subjective metrics include GPT Ratings, where the GPT model is used to score the overall quality of generated scenes. These subjective scores cover the following aspects: Object Pose (OP), which evaluates the rationality of object poses; Physical Reality (PR), which checks the physical authenticity of the scene; Semantic Consistency (SC), which verifies the semantic consistency of objects in the scene; Scene Functionality (SF), which evaluates whether the scene meets practical functional requirements; and Visual Aesthetics (VA), which measures the visual aesthetics of the scene.

Objective metrics and subjective metrics together form a comprehensive and systematic framework for performance evaluation of the IL3D dataset, which is applicable to various tasks of 3D scene generation and scene understanding. The specific calculation methods are provided in Appendix \ref{metrics}.

\subsection{Dataset Applications and Usage Scenarios}
With its high-quality 3D assets, detailed scene layouts, and rich natural language annotations, the IL3D dataset provides important support for the fields of 3D scene understanding, generation, and interaction, meeting the diverse needs of academic research and industrial applications. The scene readability and multi-level annotations of the IL3D dataset make it an ideal resource for developing and evaluating 3D scene synthesis algorithms-especially for LLM-driven scene generation. After simple supervised fine-tuning, LLMs can generate complex indoor scenes with semantic consistency and physical authenticity using this dataset.

In addition, the high-fidelity scenes provided by the dataset are compatible with multiple simulation platforms, offering key support for robot navigation and interaction tasks. Researchers can use IL3D to train models, enabling robots to perform path planning, object grasping, or scene understanding in virtual environments, thereby simulating real-world indoor interaction scenarios. The natural language annotations of IL3D also provide unique opportunities for multimodal learning tasks, allowing researchers to develop vision-language models for indoor scenes and explore cross-modal correlations between 3D scenes and natural language.

%% file: sec/4_dataset_analysis.tex
\section{Dataset Analysis}
\subsection{Room Type}
\begin{figure}[h]
    \centering
    \includegraphics[width=0.8\linewidth]{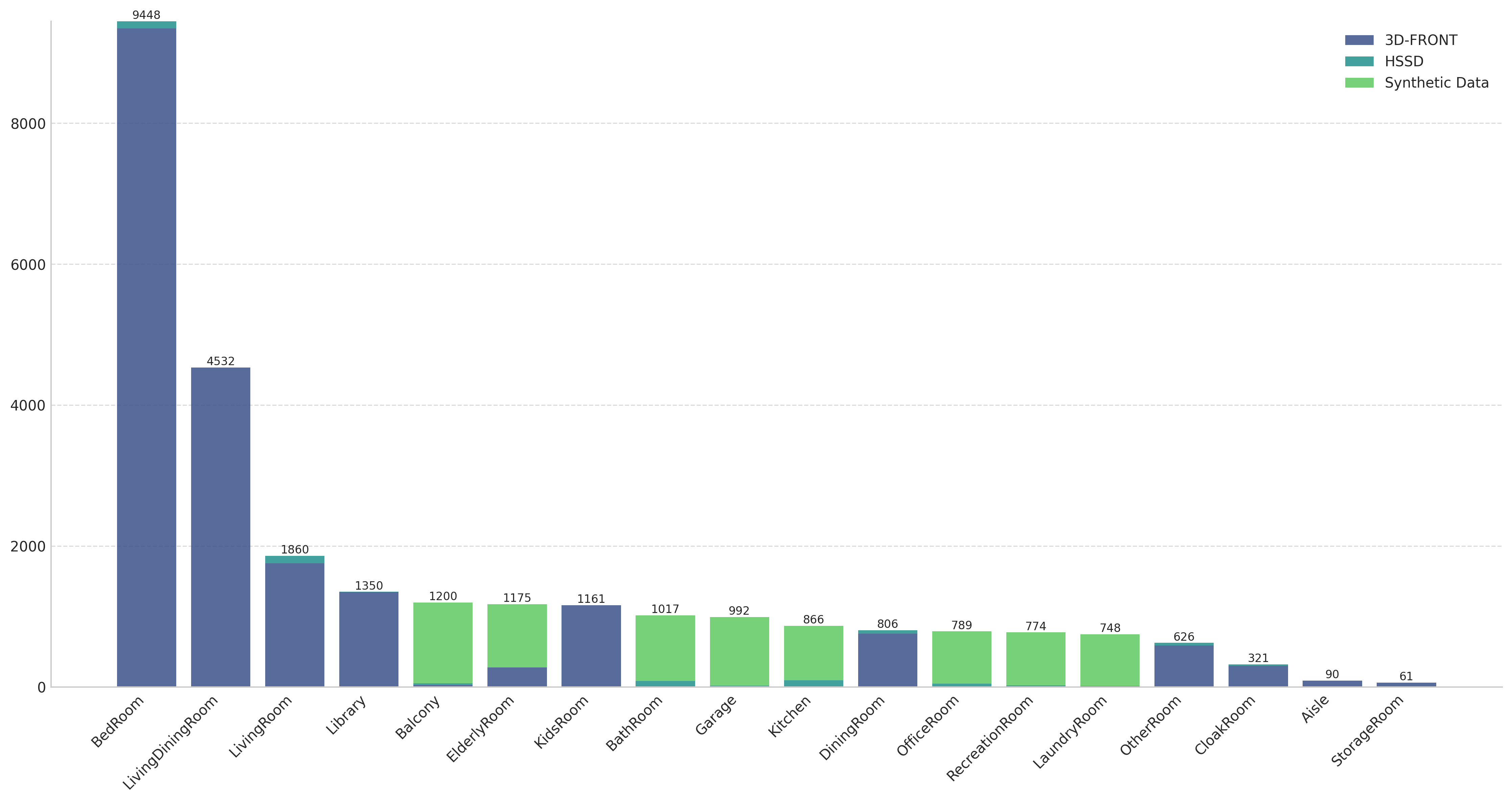}
    \caption{\textbf{Statistics of room type.} IL3D comprises 27,816 distinct rooms, consisting of 18 common room types. For existing datasets, we selectively supplemented some missing room types.}
    \label{fig: room_type}
\end{figure}
As shown in Fig. \ref{fig: room_type}, we analyzed the room type distribution of the dataset, this distribution reflects data integrated from the 3D-FRONT and HSSD datasets, supplemented by our targeted additional synthetic data. Among the room types, the bedroom category occupies a dominant position, with 9,148 samples, while living rooms and dining rooms follow closely, with 4,532 and 1,860 samples respectively. These scenes are primarily sourced from the 3D-FRONT dataset, underscoring its significant role in this field.  

Our synthetic data (highlighted in green) addresses underrepresented categories in other datasets. It enhances the diversity and balance of the dataset, thereby alleviating the lack of scenes belonging to certain specific categories in these existing datasets. The HSSD dataset, limited by its data scale, makes a relatively small contribution to scene layouts.  

Overall, the IL3D dataset contains over 27,000 samples, covering 18 room types. Among these samples, synthetic data accounts for approximately 20–30\% of the total. This not only strategically expands the coverage of indoor scene categories but also fosters the robustness of LLM-driven scene generation and other visual tasks by reducing potential biases in model training.

\subsection{Number of Objects in Rooms}
\begin{figure}
    \centering
    \includegraphics[width=0.8\linewidth]{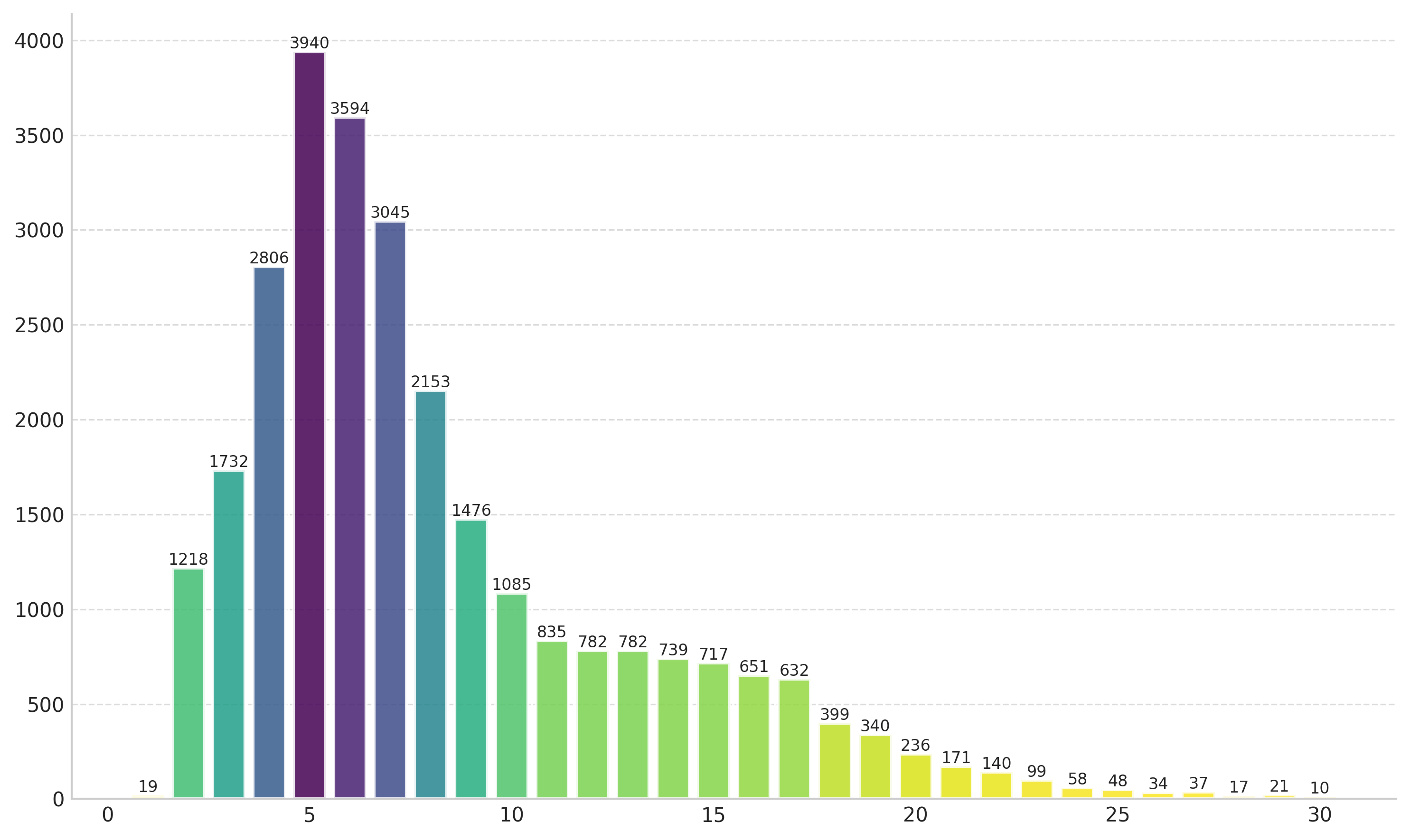}
    \caption{\textbf{Statistics of object numbers per room.} The number of objects in a room is mostly concentrated in the range of 4 to 9. When this range is exceeded, the corresponding number of rooms shows a steady downward trend as the number of objects increases.}
    \label{fig: room_obj}
\end{figure}

As shown in the Fig. \ref{fig: room_obj}, we counted the distribution of rooms with different numbers of objects. The number of objects in rooms mainly ranges from 4 to 9. Rooms containing 5 objects are the most common, with a count of 3,594; this number primarily corresponds to the most typical indoor room types, such as living rooms and bedrooms.

Starting from rooms with 10 objects, the number of corresponding rooms drops below 1,000 and decreases steadily afterward, which reflects a good balance in the layout of moderately complex rooms. Even when the number of objects exceeds 20 (with the number of rooms falling below 200), there are still a considerable number of highly complex scenes in IL3D-this highlights the comprehensiveness of our dataset.
\subsection{Room Area}
\begin{figure}
    \centering
    \includegraphics[width=\linewidth]{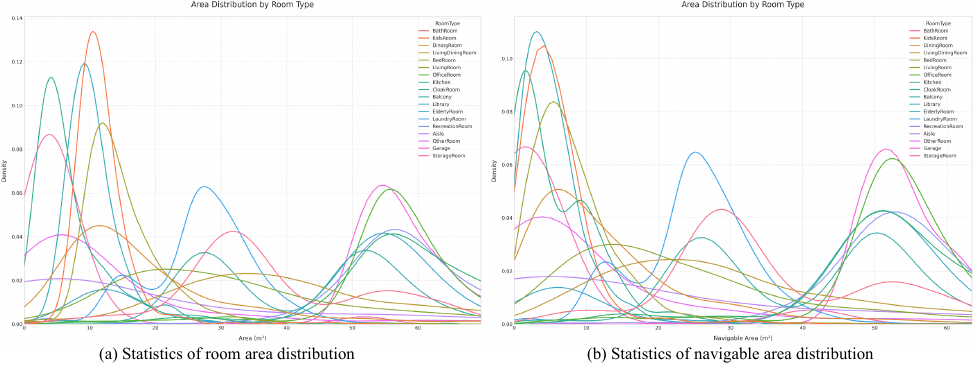}
    \caption{\textbf{Room Area Statistics.} Navigable areas generally exhibit a certain degree of deviation compared with room areas.}
    \label{fig: area}
\end{figure}

The area distribution of various room types in our indoor scene layout dataset reveals rich patterns of spatial features, as shown in Fig. \ref{fig: area}.

Fig. \ref{fig: area} (a) presents the density distribution of the total room area; most room types exhibit a multi-peak feature, reflecting the standardization of architectural design and the diversity of functional requirements. For example, the distribution of bathrooms and kitchens shows a significant peak around 5–15 square meters, embodying their compact design that prioritizes practicality; in contrast, larger spaces such as living rooms and garages have a wider distribution, with peaks exceeding 20 square meters, adapting to diverse furniture layouts and vehicle storage needs.

In comparison, Fig. \ref{fig: area} (b) shows the distribution of navigable area-a metric that accounts for obstacle-free regions used for movement. Notably, these distributions are shifted leftward compared to the total area, with reduced peak density and a narrowed range, highlighting the significant impact of fixed facilities, furniture, and built-in elements on non-navigable areas. For instance, the navigable area of kitchens and dining rooms is reduced by approximately 20–30\% compared to their total area, which may be attributed to cabinets and appliances; meanwhile, open spaces such as atriums and entertainment rooms maintain a high proportion of navigable area.

This difference underscores the value of the dataset in robotics and virtual reality applications: as a core metric for path planning and accessibility evaluation, navigable area provides critical references. In general, the total area reflects the functional attributes of a room, while the navigable area offers a more appropriate perspective for understanding human-centered spatial practicality in indoor environments.
\subsection{Object Categories}
\begin{wrapfigure}{r}{0.38\linewidth}
    \centering
    \includegraphics[width=1\linewidth]{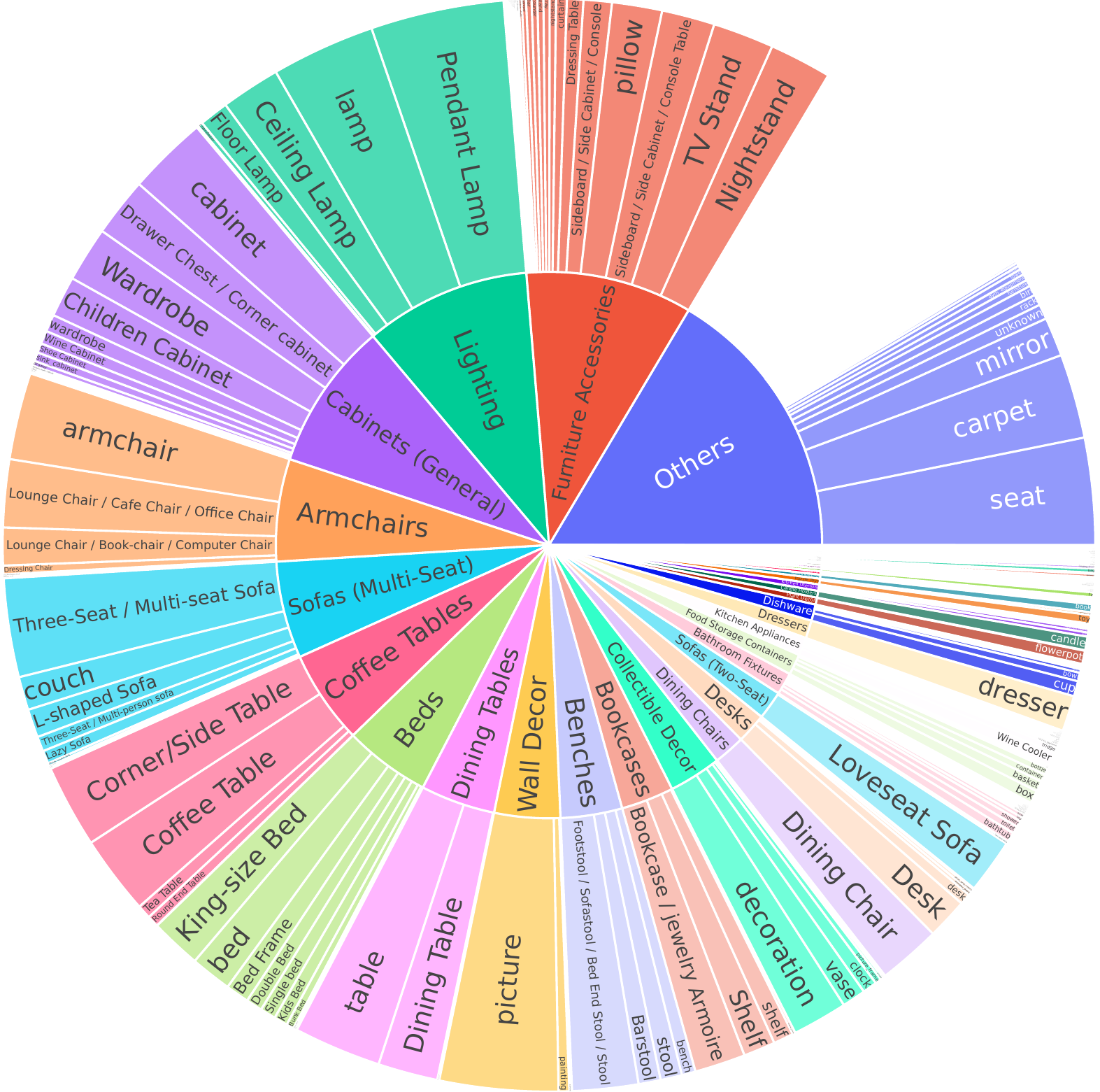}
    \caption{Statistics of object category distribution.}
    \label{fig: object_category}
\end{wrapfigure}
We counted object category distribution in the 3D asset library to understand the composition and diversity of common indoor elements. The pie chart in Fig. \ref{fig: object_category} shows the dataset’s object categories hierarchically grouped into major types (Furniture, Lighting, Accessories, Others), each split into specific subcategories. For example, Furniture subcategories like Armchairs and Multi-Seat Sofas make up a big share-key for spatial functions; Lighting includes Pendant Lamps and Table Lamps, meeting different lighting needs. This hierarchy not only clarifies the modular nature of object categories but also reveals large proportional differences. For instance, Furniture accounts for over 70\%, far more than others-due to the dataset focusing on residential and office indoor scenes. Overall, the distribution shows object categories are function-focused: high-frequency items (e.g., commonly used furniture) dominate, while low-frequency ones add realism and diversity to scenes.

%% file: sec/5_experiments.tex
\section{Experiments}
We investigated the impact of different dataset scales and scene data settings on LLM-Driven scene generation tasks, focusing on the performance of the same LLM after Supervised Fine-Tuning (SFT) under varying training data conditions.
\subsection{Experimental Setup}
The scene generation process was divided into two stages: 3D asset retrieval and scene layout generation. To study the influence of natural language annotations on the spatial reasoning ability of LLMs, we adopted a ``retrieval-then-generation" strategy-first retrieving object information in the scene based on text descriptions, then using the SFT-tuned LLM for reasoning and generation.

The test data consisted of 30 randomly generated natural language descriptions for six common room types. GPT-4o was used as the scoring model for subjective metrics. Qwen3 models with different parameter sizes were all fine-tuned via LoRA (Parameter-Efficient Fine-Tuning). Detailed experimental configurations are provided in Appendix \ref{detail_exp}.
\subsection{Experimental Results}
I-Design and HOLODECK were selected as baselines to compare the performance of our model with other LLM-Driven scene generation methods.
\subsection{Main Results}
\begin{table}[t]
    \caption{\textbf{Main experimental results.} Comparison of performance in objective and subjective metrics across I-Design, HOLODECK and Qwen3-14B (Supervised Fine-Tuning on IL3D).}\label{tab: compare_res}
    \centering
    \resizebox{1\textwidth}{!}{
    \begin{tabular}{@{}lcccccccccc@{}}
        \toprule
        \multicolumn{1}{l}{\multirow{2}{*}{Methods}} & \multicolumn{1}{c}{\multirow{2}{*}{Reference}} & \multicolumn{4}{c}{Objective Metrics} & \multicolumn{5}{c}{Subjective Metrics}\\
        \cmidrule(lr){3-6} \cmidrule(lr){7-11}
           & & \quad OOB$\downarrow$ & OOR$\downarrow$ & CLIP-Sim$\uparrow$ & GSR$\uparrow$ & OP$\uparrow$ & PR$\uparrow$ & SC$\uparrow$ & SF$\uparrow$ & VA$\uparrow$ \\
        \midrule
        I-Design & ECCV 2024 & 39.883 & 17.466 & \underline{26.406} & 37.8\% & \textbf{7.417} & \textbf{8.767} & \textbf{7.467} & \textbf{6.5} & \textbf{7.317}\\
        HOLODECK & CVPR 2024 & \textbf{1.783} & \textbf{0.113} & 25.164 & 91.7\% & 6.65 & \underline{8.2} & \underline{6.417} & \underline{5.85} & \underline{6.767}\\
        Qwen3-14B & - & \underline{16.455} & \underline{4.494} & \textbf{27.357} & \textbf{100\%} & \underline{6.683} & 7.967 & 5.783 & 5.467 & 6.533\\
        \bottomrule
  \end{tabular}
  }
\end{table}

As shown in Tab. \ref{tab: compare_res}, we compared the 3D scene generation performance of I-Design, HOLODECK, and our SFT-tuned Qwen3-14B. In terms of objective metrics, Qwen3-14B exhibited excellent performance: its Out-of-Bound (OOB: 16.455) and Object Overlap Rate (OOR: 4.494) were much lower than those of I-Design, while its CLIP-Similarity (CLIP-Sim: 27.357) and Generation Success Rate (GSR: 100\%) were higher than both baselines. This demonstrates the advantages of the SFT method in generation reliability and semantic alignment, which is attributed to the fact that one-time forward reasoning avoids the cumulative reasoning errors of agent-based methods.

In terms of subjective metrics, I-Design outperformed Qwen3-14B in Object Pose (OP), Physical Reality (PR), Semantic Consistency (SC), Scene Functionality (SF), and Visual Aesthetics (VA). HOLODECK was close to I-Design in PR but slightly inferior in other metrics, while Qwen3-14B showed weaker performance in SC and SF due to the lack of iterative optimization.

In summary, the SFT method is efficient and stable, making it suitable for rapid scene generation; agent-based methods improve subjective quality through iterative optimization, but their stability needs further enhancement.

\subsection{Ablation Study}
\subsubsection{Impact of Dataset Scale}
As shown in Tab. \ref{tab: dataset_res}, we evaluated the 3D scene generation performance of the Qwen3-1.7B model after SFT on the HSSD, 3D-FRONT, and IL3D datasets, where the dataset scales increased in the order of HSSD $<$ 3D-FRONT $<$ IL3D.

\begin{table}[t]
  \centering
  \small
  \begin{minipage}[t]{0.49\textwidth}
    \centering
    \caption{Comparison of performance in objective indicators for Supervised Fine-Tuning with Qwen3-1.7B on datasets of different scales.}
    \resizebox{1\textwidth}{!}{
    \begin{tabular}{@{}lcccc@{}}
      \toprule
        Datasets & OOB$\downarrow$ & OOR$\downarrow$ & CLIP-Sim$\uparrow$ & GSR$\uparrow$ \\
        \midrule
        \multicolumn{5}{c}{Unannotated Inference} \\
        \midrule
        HSSD & \textbf{4.039} & 36.978 & 24.918 & 98.9\% \\
        3D-FRONT & 20.577 & \underline{5.040} & \underline{26.484} & \underline{100\%} \\
        IL3D & \underline{5.684} & \textbf{4.372} & \textbf{26.498} & \textbf{100\%} \\
        \midrule
        \multicolumn{5}{c}{Annotated Inference} \\
        \midrule
        HSSD & 53.448 & 35.875 & 24.998 & 100\% \\
        3D-FRONT & \underline{27.063} & \underline{6.811} & \textbf{26.912} & \underline{100\%} \\
        IL3D & \textbf{16.505} & \textbf{5.551} & \underline{26.774} & \textbf{100\%} \\
      \bottomrule
    \end{tabular}
    }
    \label{tab: dataset_res}
  \end{minipage}
  \hfill
  \begin{minipage}[t]{0.49\textwidth}
    \centering
    \caption{Comparison of performance in objective indicators for Supervised Fine-Tuning of different Qwen3 models on the IL3D dataset.}
    \resizebox{1\textwidth}{!}{
    \begin{tabular}{@{}lcccc@{}}
      \toprule
        Models & OOB$\downarrow$ & OOR$\downarrow$ & CLIP-Sim$\uparrow$ & GSR$\uparrow$ \\
        \midrule
        \multicolumn{5}{c}{Unannotated Inference} \\
        \midrule
        Qwen3-4B & 34.560 & 6.513 & 26.498 & 100\% \\
        Qwen3-8B & \underline{24.945} & \textbf{4.107} & \underline{27.007} & \underline{97.8\%} \\
        Qwen3-14B & \textbf{17.479} & \underline{5.391} & \textbf{27.013} & \textbf{100\%} \\
        \midrule
        \multicolumn{5}{c}{Annotated Inference} \\
        \midrule
        Qwen3-4B & 27.996 & 5.809 & \underline{27.193} & 100\% \\
        Qwen3-8B & \underline{21.998} & \underline{4.695} & 26.997 & \underline{100\%} \\
        Qwen3-14B & \textbf{16.455} & \textbf{4.494} & \textbf{27.357} & \textbf{100\%} \\
      \bottomrule
    \end{tabular}
    }
    \label{tab: model_res}
  \end{minipage}
\end{table}

In the unannotated reasoning scenario, IL3D achieved the best performance in OOB (5.684) and OOR (4.372), outperforming HSSD (OOB: 4.039, OOR: 36.978) and 3D-FRONT (OOB: 20.577, OOR: 5.040). It also ranked among the top in CLIP-Sim (26.498) and GSR (100\%). These results indicate that IL3D, with its larger scale and higher diversity, exhibits excellent performance in geometric control and semantic alignment.

In the annotated reasoning scenario, HSSD’s OOB metric decreased significantly (53.448), which may be due to its small scale making it difficult to adapt to complex annotation requirements. In contrast, IL3D (OOB: 16.505) and 3D-FRONT (OOB: 27.063) still maintained relatively good performance. The GSR of all datasets was close to 100\%, and their CLIP-Sim values were slightly higher than those in the unannotated reasoning scenario.

Overall, IL3D demonstrated higher stability and superiority in both unannotated and annotated reasoning scenarios, an advantage mainly attributed to its large-scale and diverse dataset design.
\subsubsection{Impact of Natural Language Annotations}
As shown in Tab. \ref{tab: dataset_res} and Tab. \ref{tab: model_res}, we evaluated the performance of the Qwen3 model series (with parameter sizes of 1.7B, 4B, 8B, and 14B) on the IL3D dataset under both unannotated and annotated reasoning conditions.

In the unannotated reasoning scenario, Qwen3-1.7B exhibited good boundary control, with an OOB value of 5.684, an OOR value of 4.372, a CLIP-Sim value of 26.498, and a GSR of 100\%, though its semantic similarity was relatively low. Qwen3-4B showed a tendency for boundary overflow, as its OOB value rose to 34.560, with an OOR of 6.513, a CLIP-Sim of 27.066, and a GSR that dropped to 94.4\%, indicating that medium-sized models are prone to such boundary issues. Qwen3-8B and Qwen3-14B, as large-sized models, demonstrated better generation stability: their OOB values decreased to 24.945 and 17.479 respectively, their OOR values were 4.107 and 5.391, their CLIP-Sim values remained stable at 27.007 and 27.013, and their GSR values rebounded to 97.8\% and 100\%.

Annotated reasoning significantly improved the performance of all models, with the GSR of each model reaching 100\%. For Qwen3-1.7B, the OOB value increased to 16.505, while the OOB values of the other three models (4B, 8B, 14B) decreased to 27.996, 21.998, and 16.455 respectively. The OOR value of Qwen3-1.7B rose to 5.551, whereas the OOR values of the other three models dropped to 5.809, 4.695, and 4.494. The CLIP-Sim values of all models fluctuated between 26.774 and 27.357, which indicates that natural language annotations mainly play a role in optimizing geometric consistency. These results further show that annotated reasoning effectively alleviates the boundary overflow problem in medium-sized and large-sized models.

In summary, the experiment revealed a trade-off between model scale and generation performance: small-sized models (e.g., Qwen3-1.7B) have high efficiency in boundary control but limited ability to capture semantic information; large-sized models (e.g., Qwen3-8B, Qwen3-14B) can optimize OOB and OOR metrics with the assistance of annotations, but their improvement in CLIP-Sim (semantic similarity) remains limited.

%% file: sec/6_conclusion.tex
\section{Conclusions}
This study constructs the IL3D dataset and applies it to LLM-Driven indoor scene generation, establishing a new benchmark in the field of 3D scene generation. IL3D not only provides diverse indoor layouts and high-precision object assets but also adapts well to visual tasks such as 3D scene generation and editing, serving as a powerful tool for complex visual tasks and embodied intelligence research. Experiments verify its significant advantages in improving model performance and adaptability. Despite its limitation in scene-level relationship descriptions, IL3D’s openness and extensibility lay a foundation for future exploration of deeper scene understanding.

\section{Acknowledgments}
We express our gratitude to the members of the University of Science and Technology of China and Songying Technology for their valuable feedback on this project. This work was primarily supported by resources provided by Songying Technology.

%% file: sec/7_appendix.tex
\clearpage
\section{Appendix}
\subsection{LLM Usage Statement}
In the process of experimental implementation and paper writing of this study, the usage of large language models (LLMs) is stated as follows:

In the experimental section, given that this work focuses on the LLM-Driven indoor scene generation task, we adopted Qwen3 series models (including versions with 1.7B, 4B, 8B, and 14B parameters) for supervised fine-tuning (SFT) to verify the effectiveness of the proposed IL3D dataset and related methods. We further combined the LoRA (Low-Rank Adaptation) parameter-efficient fine-tuning technique to optimize the model’s spatial reasoning capability, which supported the implementation of core experimental links such as scene generation performance evaluation and ablation studies.

During the paper writing phase, we used LLM tools for manuscript polishing, mainly including optimizing the accuracy of academic expressions and organizing sentence logic to conform to the writing norms of scientific research papers. However, the core ideas, research design, experimental data and result analysis, conclusion derivation, and other key contents of the paper were independently completed by the authors. We strictly ensured the originality of all academic viewpoints and the authenticity of data, in full compliance with academic research ethics and normative requirements.

\subsection{Experimental Details}
\label{detail_exp}
\subsubsection{Experiment Configuration}
\begin{table}[htbp]
  \centering
  \caption{Training configuration for different Qwen3 parameter sizes}
  \label{tab4}
  \begin{tabular}{ccccc}
    \toprule
    Config        & Qwen3-1.7B   & Qwen3-4B     & Qwen3-8B     & Qwen3-14B   \\
    \midrule
    Optimizer     & AdamW        & AdamW        & AdamW        & AdamW       \\
    Learning Rate & 1e-4         & 5e-5         & 5e-5         & 5e-5        \\
    LoRA Rank     & 8            & 8            & 8            & 8           \\
    LoRA Alpha    & 32           & 32           & 16           & 16          \\
    Target Models & all-linear   & all-linear   & all-linear   & all-linear  \\
    Warmup Ratio  & 0.05         & 0.05         & 0.05         & 0.05        \\
    \midrule
    Batch Size    & 2            & 2            & 2            & 2           \\
    Max Length    & 4096         & 4096         & 4096         & 4096        \\
    \midrule
    GPU device    & NVIDIA L40S  & NVIDIA L40S  & NVIDIA L40S  & NVIDIA L40S \\
    \midrule
    Training Time & $\sim$ 4h & $\sim$ 5h & $\sim$ 7h & $\sim$ 12h \\
    \bottomrule
  \end{tabular}
\end{table}
As shown in Tab. \ref{tab4}, we provide the detailed configuration information of the experiment, including parameter settings, hardware conditions, and training duration.

\subsubsection{3D Asset Retrieval}
\label{3d_retrieval}
\begin{figure}
    \centering
    \includegraphics[width=1\linewidth]{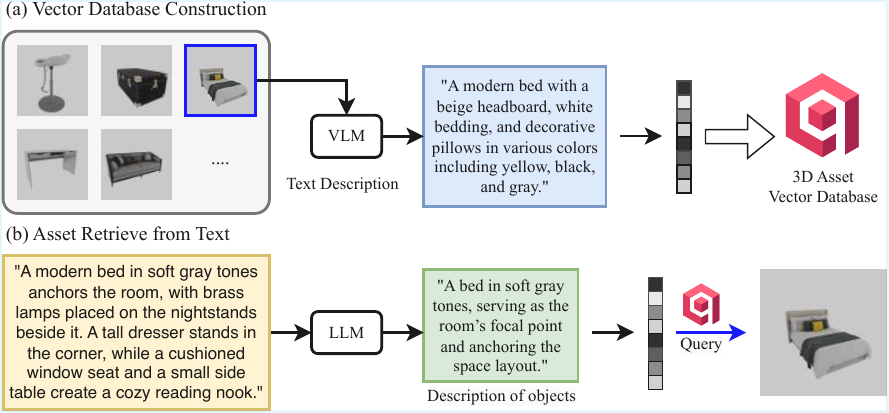}
    \caption{Asset Retrieval. (a) A 3D asset vector database for object descriptions is constructed based on Qdrant. (b) When generating scenes, LLMs are used to extract object information from text descriptions and retrieve relevant assets.}
    \label{fig: retrieve}
\end{figure}
As shown in the Fig. \ref{fig: retrieve}, this study implements 3D asset retrieval based on the instance-level annotations of the IL3D dataset, primarily employing Qdrant\footnote{https://qdrant.tech/} to construct a vector database of text descriptions. This vector database maps each 3D asset to a vector representation corresponding to its instance-level annotations (e.g., descriptions of object type, appearance, and material), laying the data foundation for subsequent similarity-based retrieval.  

During the text-driven scene generation process, a LLM is first used to extract specific descriptions of the objects required in the scene from the text description of the target scene; these descriptions cover core information such as object category, key features, and functional attributes. The extracted object descriptions are then converted into vector representations, and the vector database is queried based on this vector to find the text description with the highest semantic similarity, thereby retrieving the corresponding 3D asset. This ensures that the selected asset aligns with the semantic intent of the scene’s text description, providing a reliable asset foundation for subsequent scene layout generation.

\subsection{Supplementary Related Work} 
Image-guided scene generation focuses on extracting 3D spatial information from input images, typically implemented via end-to-end generative models or frameworks composed of multiple visual models. This direction enables scene reconstruction or synthesis directly from 2D image cues, providing critical support for tasks such as virtual reality, robotic perception, and 3D content creation.  

CAST~\cite{yao2025cast} proposes a component-aligned 3D scene reconstruction method from a single RGB image. It first extracts object-level 2D segmentation and relative depth information, then leverages foundation models to infer 3D geometry and semantic relationships. This approach achieves high-fidelity scene recovery across diverse indoor and outdoor settings, laying a foundation for applications in virtual reality and robotics that require accurate 3D scene representations. Gen3DSR~\cite{ardelean2024gen3dsr} is a generalizable 3D scene reconstruction framework that follows a ``divide-and-conquer strategy''. It first processes the scene globally to extract depth and semantic information, then performs hierarchical optimization for object-level reconstruction. By generating compositional 3D scenes from single-view images, it demonstrates strong generalization capabilities to unseen environments, addressing the challenge of limited adaptability in traditional reconstruction methods.   

DiffScene~\cite{tang2024diffuscene} adopts a diffusion denoising model for the synthesis of generative indoor scenes. Through a scene configuration denoising mechanism, it generates diverse and photorealistic 3D indoor environments from a text prompt describing partial scene configurations. It supports key tasks including scene completion, object arrangement, and text-conditioned synthesis, and uses an unordered object set representation to enhance generation flexibility and applicability to downstream tasks. MIDI~\cite{huang2025midi} extends a pre-trained image-to-3D object generation framework via a multi-instance diffusion model. It generates compositional 3D scenes from a single image, capable of handling the geometry and texture of multiple instances simultaneously in a single feed-forward pass. This efficiency makes it well-suited for rapid indoor layout synthesis, where multiple objects need to be integrated coherently. SceneGen~\cite{meng2025scenegen} is a single-image 3D scene generation model that synthesizes geometry, texture, and relative poses in a single feed-forward pass, reportedly the first framework to achieve this capability. By streamlining the generation process, it enables fast construction of photorealistic indoor environments, reducing the computational overhead of multistep scene building. PhyScene~\cite{yang2024physcene} emphasizes the generation of physically interactive 3D scenes, designed specifically for embodied AI tasks. It creates indoor environments with realistic layouts and rich interactivity by incorporating physical constraints on object placement and considering articulated object interactions. This focus on physical plausibility ensures that the generated scenes can support practical embodied interactions (e.g., robot object manipulation), bridging the gap between synthetic scenes and real-world usability.

\subsection{Supplementary Data Statistics}
\subsubsection{Physical Scale of the Asset Library}
\begin{figure}[t]
    \centering
    \includegraphics[width=\linewidth]{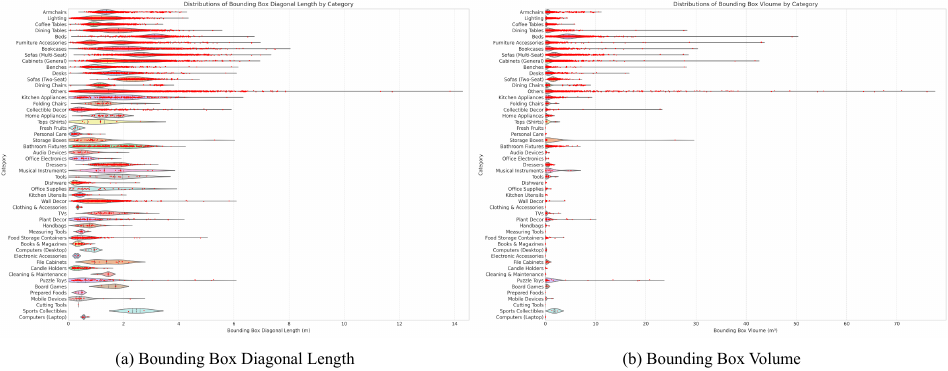}
    \caption{Statistical results of the physical scale distribution of assets.}
    \label{fig: obj_scale}
\end{figure}
This section presents statistical results on the physical dimensions of 70 object categories in the 3D asset library, focusing on two key metrics: the diagonal length of the bounding box and the volume of the bounding box.

As shown in Fig. \ref{fig: obj_scale} (a), for the distribution of the diagonal length of the bounding box (unit: meters), the distribution plot shows the diagonal length of the instances in each category. Small categories, such as fresh fruits, French fries and food measuring tools, exhibit narrow distributions, indicating their consistent and compact dimensions, which are typical of desktop or handheld items. In contrast, large furniture categories (e.g., sofas, beds, and cabinets) display wide distribution ranges, reflecting size variations from compact to spacious designs. This characteristic is critical for 3D scene understanding algorithms, as it enables them to handle spatial layouts of multiple scales.

As shown in Fig. \ref{fig: obj_scale} (b), for the volume distribution of the bounding box, all 70 categories show a similar trend of variation, and their order of classification is consistent with that of the diagonal length of the bounding box, highlighting the correlation between volume distribution and linear dimensions in 3D assets. Categories with high morphological variability (e.g., office chairs, storage racks, and countertops) exhibit broad volume distributions, which reflects the impact of modular or customizable designs on occlusion and interaction modeling. In contrast, categories such as beverages, computers, and musical instruments have narrow volume distributions, emphasizing their standardized dimensions in the real world. This standardization facilitates accurate semantic segmentation and functional prediction in indoor 3D perception systems.

\subsubsection{Quantity of Different Asset Categories}
\begin{figure}
    \centering
    \includegraphics[width=\linewidth]{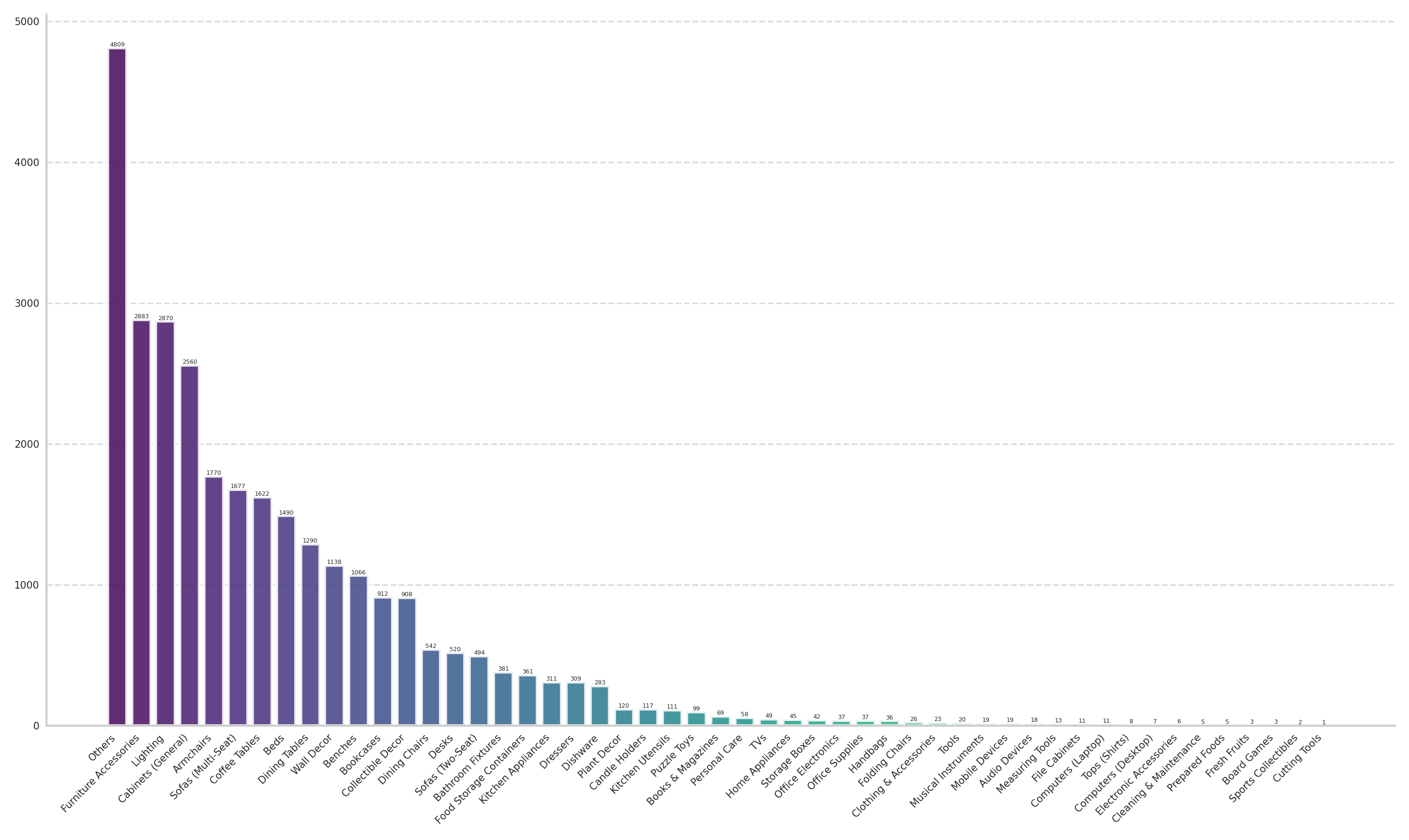}
    \caption{Statistical results of different asset categories.}
    \label{fig: obj_num}
\end{figure}
The bar chart (shown in Fig. \ref{fig: obj_num}) presents the quantity distribution of various object categories, sorted in descending order of quantity, highlighting the dominant status of certain objects in the dataset. For example, the ``Others" category-comprising daily miscellaneous items such as seats and carpets - ranks first with more than 4,000 instances, reflecting the universality and functional necessity of these objects in daily indoor spaces. It is followed by furniture categories such as sofas and tables, whose quantities gradually decrease from common to rare types until approaching zero.

This long-tailed distribution indicates that the dataset captures the frequency pattern of objects in the real world: a small number of high-frequency objects (e.g., seating and storage items) dominate, while a large number of low-frequency objects (e.g., specific decorative items) contribute to diversity. This characteristic embodies the hierarchical and personalized features of indoor design.

\subsubsection{Evaluation Metrics}
\label{metrics}
\textbf{Out-of-Bound (OOB).} The Out-of-Bound (OOB) metric evaluates whether objects in a generated indoor scene exceed room boundaries (e.g., walls, floors, or ceilings). It ensures the physical plausibility of scenes, preventing irrational cases like objects floating or penetrating boundaries-an essential requirement for applications such as virtual reality, robotic simulation, and architectural design. A lower OOB value implies the generated scene better adheres to spatial constraints, thus enhancing realism and usability.
\begin{equation}
    \text{OOB} = \frac{1}{N} \sum_{i=1}^{N} \mathbb{I}(V_{i,\text{out}} > 0)
\end{equation}
where N is the total number of objects in the scene, $V_{i,\text{out}}$ denotes the volume of the i-th object that exceeds room boundaries, and $\mathbb{I}(\cdot)$ is an indicator function (taking 1 if the condition holds, 0 otherwise).

\textbf{Object Overlap Rate (OOR).} The Object Overlap Rate (OOR) quantifies the overlap degree between objects in a generated scene. In indoor scenarios, object overlaps (e.g., a chair penetrating a table) create unnatural layouts and reduce realism. OOR measures such collisions or overlaps to optimize generative models for physically reasonable object placement.
\begin{equation}
    \text{OOR} = 
        \begin{cases} 
        \frac{\sum_{i=1}^{N} \sum_{j=i+1}^{N} V_{i,j}^{\text{inter}}}{\sum_{i=1}^{N} V_i} & \text{if } \sum_{i=1}^{N} V_i \geq 10^{-9}, \\
        0 & \text{otherwise}
        \end{cases}
\end{equation}
where $V_{i,·}$ is the bounding box volume of the i-th object, and $V_{i,j}^{\text{inter}}$ is the intersecting volume of the i-th and j-th objects’ bounding boxes (only i<j is computed to avoid duplicate counting of object pairs). If the total volume of all objects is extremely small, the OOR is set to 0 to avoid division errors by zero.

\textbf{Generation Success Rate (GSR).} The Generation Success Rate (GSR) measures the proportion of successfully generated scenes, i.e., whether the model can output valid scene results (as opposed to failure cases such as incomplete generation or termination due to errors). GSR reflects the generative model’s ability to stably produce scenes, which is fundamental for evaluating the reliability of the generation process itself.
\begin{equation}
    \text{GSR} = \frac{1}{M} \sum_{m=1}^{M} S_m
\end{equation}
where $S_m=\mathbb{I}(\cdot)$(the m-th scene is successfully generated) (taking 1 if the scene is generated successfully, 0 otherwise), and M is the total number of generation attempts.

\textbf{CLIP-Similarity (CLIP-Sim).} CLIP-Similarity (CLIP-Sim) uses the CLIP model\footnote{https://huggingface.co/openai/clip-vit-base-patch32} (Contrastive Language-Image Pretraining) to measure semantic similarity between a generated scene and a reference scene/text description. It assesses whether the generated scene aligns with expectations visually or semantically, which is vital for tasks requiring alignment with user intentions or reference images.

\begin{figure}
    \centering
    \includegraphics[width=0.8\linewidth]{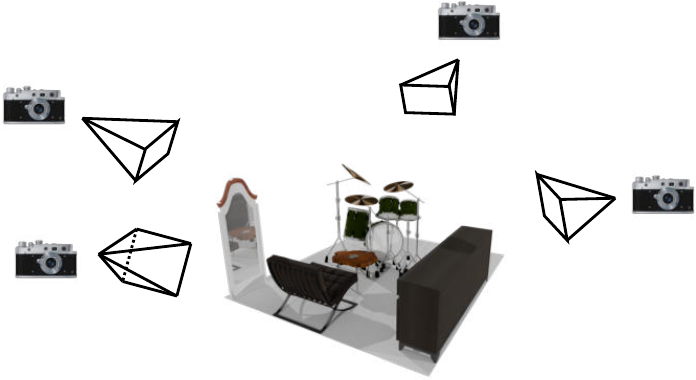}
    \caption{To prevent occlusion between objects from affecting the evaluation of CLIP-Sim, we render one image every 90 degrees of rotation along the outer side of the scene, resulting in a total of 4 rendered images. Finally, the maximum CLIP-Sim value is used as the score for this scene.}
    \label{fig: clip}
\end{figure}

For the m-th generated scene, we first calculate its CLIP similarity with the text description from each of the four predefined viewpoints (v=1,2,3,4). The final CLIP similarity for the scene is defined as the maximum value across these four viewpoints:

\begin{equation}
    \text{CLIP-Sim}_m = \max_{v \in \{1,2,3,4\}} \left( \frac{\mathbf{f}_{\text{view},m,v} \cdot \mathbf{f}_{\text{text}}}{\|\mathbf{f}_{\text{view},m,v}\| \cdot \|\mathbf{f}_{\text{text}}\|} \right)
\end{equation}

where $f_{view,m,v}$ is the feature vector of the m-th scene from the v-th viewpoint, $f_{text}$ is the feature vector of the text description, $\cdot$ denotes the vector dot product, and $\left \| \cdot \right \|$ denotes the L2 norm of a vector.

\textbf{Subjective Metrics.} As shown in Fig. \ref{fig:prompt-3}, we provided the prompt words for subjective scoring.

\subsubsection{Text Readability of USD}
\begin{figure}
    \centering
    \includegraphics[width=1\linewidth]{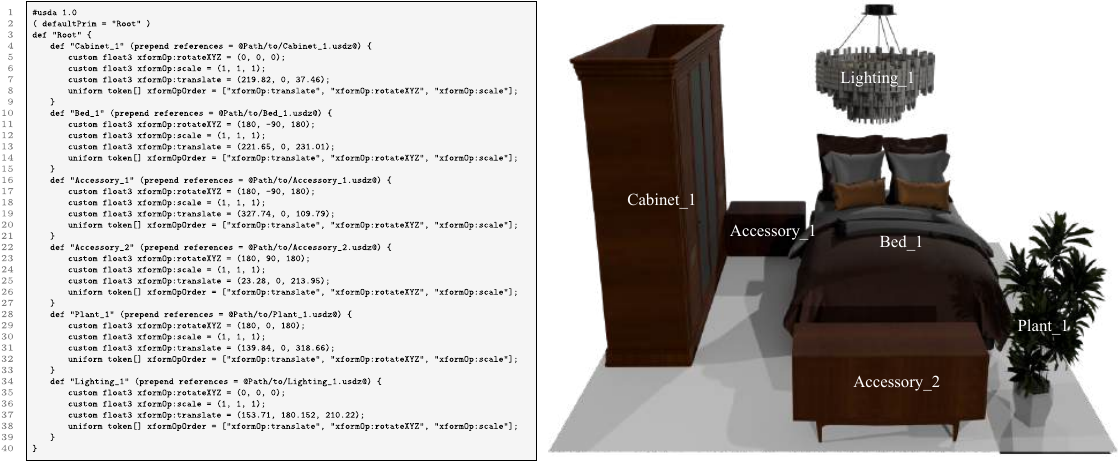}
    \caption{\textbf{Text readability.} Scenes in USD format feature text readability, which facilitates LLM-driven methods to directly generate and edit 3D scenes.}
    \label{fig:placeholder}
\end{figure}

The text representation of Universal Scene Description\footnote{https://www.nvidia.cn/omniverse/usd/} (USD) organizes objects and their transformations (e.g., translation and scaling attributes) in indoor scenes via an ASCII-based structure, ensuring high readability. This feature enables direct editing of object positions and dimensions, facilitating debugging and collaborative design workflows. The corresponding 3D visualization aligns precisely with the text data, demonstrating how the structured format of USD supports the accuracy of spatial mapping and rapid prototyping in the development of 3D scenes, particularly in complex indoor environments.

Beyond boosting the efficiency of manual editing and debugging, this readability also forms a foundation for LLM-driven 3D scene generation. By parsing USD’s hierarchical text, LLMs can interpret object relationships and transformation parameters to generate or optimize scene layouts. For example, guided by natural language instructions, LLMs can dynamically adjust object positions or introduce new elements, leveraging USD’s structure to enable automated scene design.

\begin{wrapfigure}[20]{r}{0.6\linewidth}
    \centering
    \includegraphics[width=\linewidth]{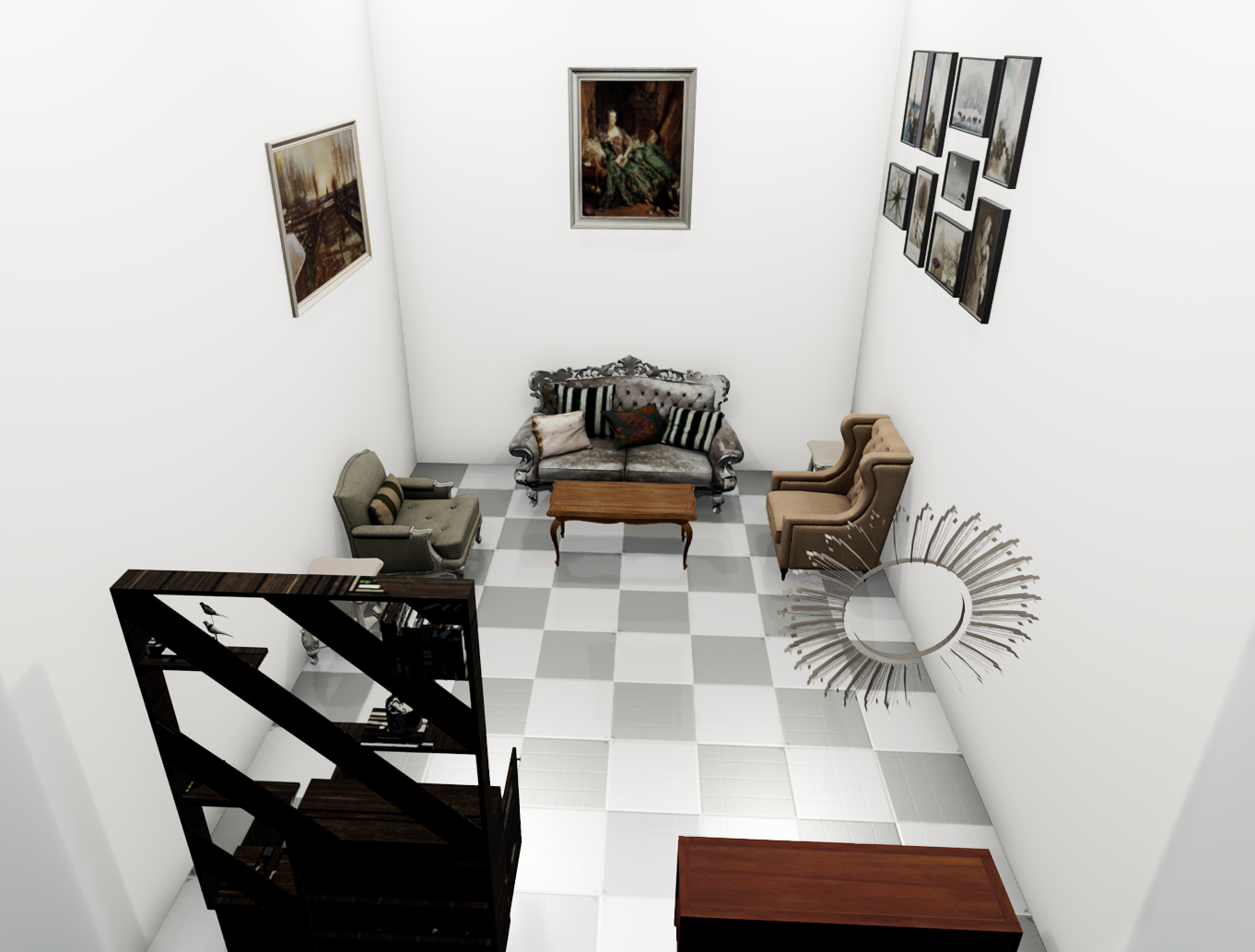}
    \caption{Importing the scenes in IL3D into simulation software can support experiments related to embodied AI.}
    \label{fig:agent}
\end{wrapfigure}

As shown in Fig. \ref{fig:agent}, IL3D is adaptable to a range of LLM-driven visual tasks in complex indoor scenes, while also being compatible with multiple simulation platforms. It supports embodiment-related simulation experiments, accelerates model development, enables data-driven generation pipelines, and preserves text readability for further refinement.

\subsection{Multimodal Data Export}
\begin{figure}
    \centering
    \includegraphics[width=0.8\linewidth]{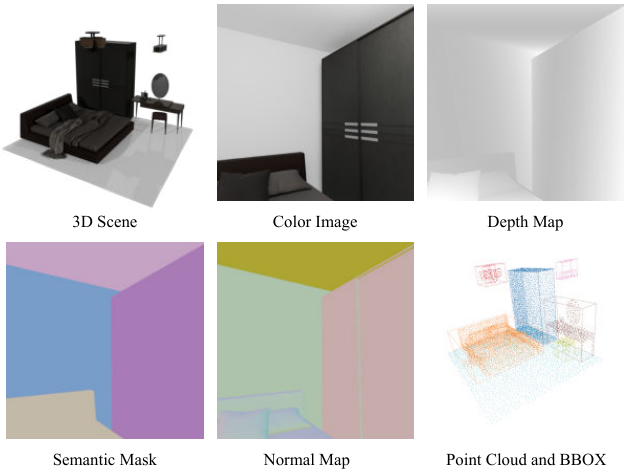}
    \caption{IL3D can export multiple types of data.}
    \label{fig: mm_export}
\end{figure}
As shown in the Fig. \ref{fig: mm_export}, the IL3D dataset supports flexible multimodal data export capabilities, enabling users to extract data tailored to specific task requirements.

In terms of 3D data, it covers core types such as semantic point clouds and 3D bounding boxes of objects. These 3D data retain fine-grained scene geometric and semantic information, providing direct input for tasks like 3D scene segmentation, object pose estimation, and spatial relationship analysis. For 2D multimodal data, users can export multi-view color images, depth maps, normal maps, and semantic masks by flexibly configuring camera parameters (e.g., viewing angle, resolution, and spatial position). This multi-view setting ensures comprehensive coverage of scene details, avoiding information loss caused by single-view occlusion.

The exported multimodal data adheres to standard data formats, enabling seamless integration with various visual task pipelines, such as using color images and semantic masks for 2D instance segmentation, or combining depth maps with 3D bounding boxes for cross-modal scene understanding. This adaptability makes IL3D a versatile resource for both academic research and industrial application development.

\subsection{Analysis and Discussion}
\subsubsection{Timing of Object Retrieval}
Existing methods mainly adopt two strategies for object retrieval in scene generation: post-retrieval and pre-retrieval. Our experiments demonstrate that the choice of retrieval strategy significantly affects the LLM's reasoning performance for indoor scene generation.

Specifically, post-retrieval, which involves the model directly predicting a general object distribution, is highly prone to inter-object penetration and objects exceeding room bounding boxes due to the lack of specific object dimensions.

In contrast, pre-retrieval provides the LLM with detailed information (e.g., scales and descriptions) of objects during reasoning, thereby improving the quality of generated scenes. However, for SFT-based methods, this approach requires a certain level of complex reasoning capability. Consequently, in models with smaller parameter sizes, introducing pre-retrieval may lead to a degradation in some metrics.

\subsubsection{Can LLMs Perceive Fine-Grained Object Shapes?}
\begin{figure}[t]
    \centering
    \includegraphics[width=1\linewidth]{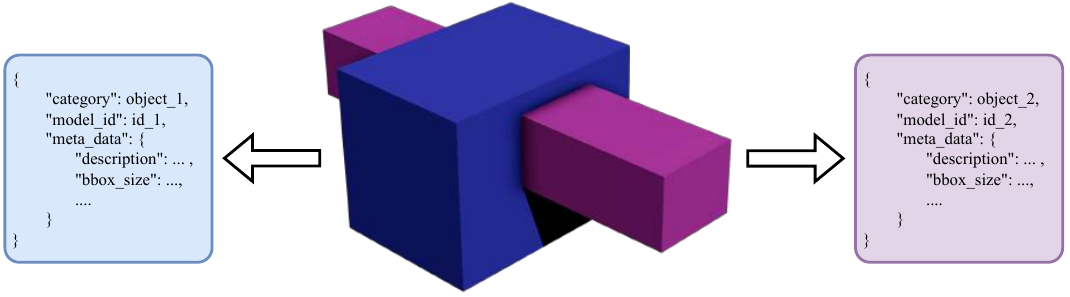}
    \caption{LLM read the information about objects' bounding boxes and their structural descriptions, and their understanding of collisions is also at the bounding box level.}
    \label{fig: bbox}
\end{figure}

As shown in Fig. \ref{fig: bbox}, objects processed by LLM-Driven scene generation methods are represented as annotated bounding boxes. Since LLMs cannot perceive the fine-grained surface shapes of objects, they can only be instructed via prompts to minimize bounding box overlap, but this often leads to physically implausible scenes, such as irrational surface contacts between objects and floating objects.

Existing methods (e.g., InstructScene~\cite{lin2024instructscene}, MetaScenes~\cite{yu2025metascenes}) address this issue by sampling point clouds from object surfaces, using pre-trained point cloud encoders to extract 3D shape representations, and inputting these representations into LLMs for reasoning after projection via linear layers; these approaches enable LLMs to perceive fine-grained shapes. Other methods artificially create penetration samples and use Direct Preference Optimization (DPO) to train LLMs to avoid inter-object penetration. While these methods partially resolve the aforementioned problems, abnormal surface contacts between objects remain difficult to completely eliminate.

We argue that a fundamental solution requires designing a method that quantifies inter-object penetration reasonably while ensuring differentiability, and integrating this physical prior into the training of end-to-end generative models. For penetration quantification, mesh-based methods incur excessive computational costs; sampling point clouds with normal directions~\cite{peng2021shape}, however, can balance computational efficiency while preserving topological structures, making it a potentially feasible solution.

\subsubsection{Agent-Based vs. SFT-Based Methods}
Agent-based methods leverage multi-agent systems to repeatedly invoke models for observing and adjusting scene data. Unlike the aforementioned point cloud encoder-based methods, agent-based approaches can render scene images and use vision-language models (VLMs) to observe spatial relationships between objects, enabling high-quality scene generation. Nevertheless, this method has two potential issues:

\begin{itemize}
    \item \textbf{Unstable generation}: Agents may continuously adjust object poses without meeting the termination condition for adjustments, leading to prolonged generation tasks that fail to produce valid scenes and waste significant computational resources.
    \item \textbf{Limited VLM observation}: VLMs struggle to detect objects with abnormal scaling (e.g., an excessively large object enclosing the entire scene and camera). In such cases, VLMs cannot identify the anomaly from rendered images or adjust the object’s pose accordingly.
\end{itemize}

SFT-based methods fine-tune LLMs to learn spatial reasoning capabilities from training data. During reasoning, the model directly accesses object information and performs a limited number of reasoning steps. Although the generated quality is slightly inferior, this approach is more robust and efficient overall.

\subsubsection{Limitations}
While the IL3D dataset provides rich instance-level natural language annotations for 3D scene generation and multimodal learning, it has a limitation: the lack of detailed scene-level descriptions of relationships between objects. In contrast, methods like InstructScene capture spatial and functional associations between objects (e.g., “a lamp on the table” or “a chair near the window”) via explicit semantic relationship graphs. IL3D’s annotations, however, focus primarily on the attributes and descriptions of individual objects, failing to fully express semantic and topological relationships between objects.

This limitation may restrict the model’s performance in complex scene understanding and reasoning tasks-especially in applications like scene editing that require deep contextual associations. We plan to introduce scene-level spatial-semantic relationship annotations in future work to further enhance IL3D’s applicability in complex 3D scene generation and interaction tasks.
\subsection{Visualization and Prompts}
We present the 3D scene visualization generated by the LLM in the experiment, as well as the designed prompts.

\begin{figure}[h]
    \centering
    \includegraphics[width=1\linewidth]{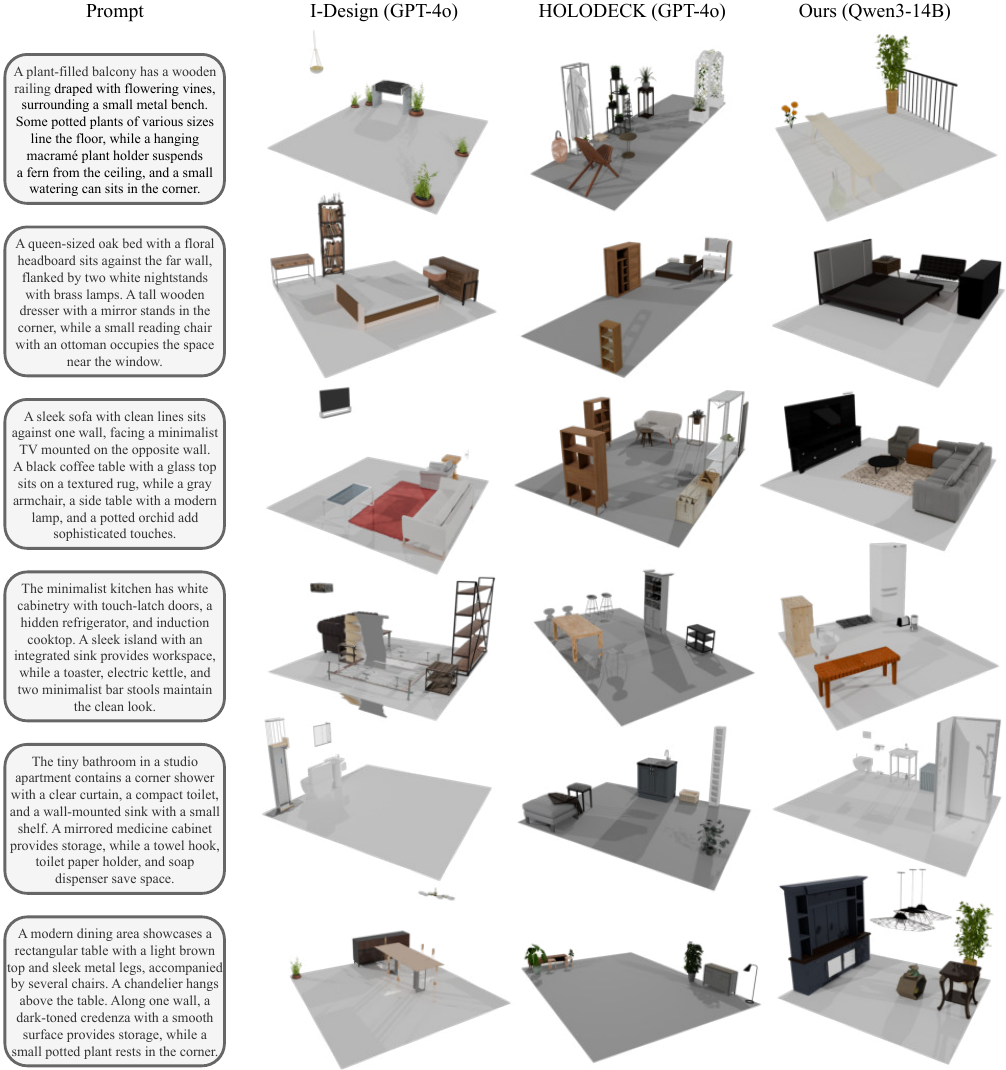}
    \caption{Generated layout comparison across models on I-Design, HOLODECK and ours.}
    \label{fig1}
\end{figure}

\begin{figure*}[h]
    \centering
    \begin{center}
    \begin{tcolorbox} [top=2pt,bottom=2pt, width=\linewidth, boxrule=1pt]
    {\footnotesize {\fontfamily{zi4}\selectfont
\textbf{Role}\\
Your task is to arrange some objects within a given \{room\_type\} effectively. Follow these guidance to complete your design:\\
\\
\textbf{Rules}\\
(1) Extract the [Objects] and [Bounding Box Size] from the object information.\\
(2) Analyze the spatial relationships among [Objects] within the specified [Room Type]. Pay special attention to **avoiding overlap** and **consider other spatial factors like accessibility and aesthetics**.\\
(3) Determine and design the precise location of all [Objects] ensuring that their bounding boxes do not overlap and that the layout is functional and visually appealing.\\
(4) I prefer objects to be placed at the edge (the most important constraint) of the room if possible which makes the room look more spacious.\\
(5) Objects usually need to be aligned in some way (such as parallel or perpendicular to the walls) and **must not extend beyond the floor area**.\\
(6) Chairs must be placed near to the table/desk and face to the table/desk.\\
(7) Before specifying the detailed positions of each object, first think about their general arrangement and relative spatial relationships:\\
    a) Which objects need the most space or have fixed positions (like beds, wardrobes)\\
    b) Which objects need to be grouped together (like nightstands with bed)\\
    c) Traffic flow and accessibility considerations.\\
\\
\textbf{Object Information}\\
\{object\_information\}\\
*Note: bbox format is [length, width, height] in meters*\\
\\
\textbf{Response Format}\\
First design the vertices of the floor, then report the 3D spatial coordinates and rotation angles of each object in JSON format, as follows:\\
\{\\
`Floor': \{`xyz': [[8.0, 0, 6.76], [8.0, 0, 0.0], [0.0, 0, 6.76], [0.0, 0, 0.0]]\},\\
`Coffee Tables': [\{`position': [1.62, 0.0, 2.29], `rotation': [180, 90, 180]\}],\\
`Benches': [\{`position': [1.72, 0.0, 3.66], `rotation': [0, 0, 0]\},
\{`position': [1.63, 0.0, 0.9], `rotation': [0, 0, 0]\}]\\
\}\\
\\
Important Notes about Coordinate System:\\
- Y-axis points upward (y=0 is floor level)\\
- X-axis runs along the room's length from west to east\\
- Z-axis runs along the room's width from south to north\\
- All coordinates are in meters\\
- Output nothing but the JSON (No preamble, no explanation, no additional text of any kind)
    }
    \par}
    \end{tcolorbox}
    \end{center}
    \caption{The prompts used for scene generation during the training and inference of Qwen3.}
    \label{fig:prompt-1}
    \vspace{-.3cm}
\end{figure*}

\begin{figure*}[h]
    \centering
    \begin{center}
    \begin{tcolorbox} [top=2pt,bottom=2pt, width=\linewidth, boxrule=1pt]
    {\footnotesize {\fontfamily{zi4}\selectfont
\textbf{Rules}\\
(1) Extract the room type from the description (e.g., BedRoom, LivingRoom, Kitchen, etc.).\\
(2) Identify all mentioned objects and their basic descriptions exactly as described.\\
(3) If the description mentions multiple instances of the same object, maintain the exact count.\\
(4) Don't omit the information about the objects in the description.\\
(5) When encountering quantity words (e.g., six, two, three, multiple) describing objects, split them into individual objects equal to the quantity.\\
Use the singular form of the object name, and apply the same description to each.\\
For example, "six chairs with blue upholstery" should be split into six separate objects: [{{"name": "Chair", "description": "A chair with blue upholstery"}}, ...] (repeated six times).\\
(6) Ensure quantity words are not included in the object name. Focus on the core object name in singular form (e.g., use "Chair" instead of "six chairs" or "Chairs").\\
\\
\textbf{Room Description}\\
\{room\_description\}\\
\\
\textbf{Response Format}\\
Report the result of the room\_type and dictionaries for each object in JSON format, as follows:\\
\{\\
`room\_type': `LivingRoom',\\
`objects': [\{`name': `Sofa', `A dark green upholstered ottoman with a cushioned lid and decorative brass nailhead trim along its edges.'\},\\
\{`name': `Armchair', `description': `A armchair with a sleek design, featuring a cushioned seat and backrest supported by thin metal legs.'\},\\
\{`name': `Armchair', `description': `A armchair featuring a curved backrest and armrests. It has a dark green upholstery and thin metal legs.'\},\\
\{`name': `Coffee Table', `description': `A modern coffee table with a round wooden top and three sleek legs that taper towards the bottom.'\}]\\
\}\\
\\
Important Notes about response format:\\
- Output nothing but the JSON. No preamble, no explanation, no additional text of any kind
    }
    \par}
    \end{tcolorbox}
    \end{center}
    \caption{The prompt used for extracting in-scene information during scene generation.}
    \label{fig:prompt-2}
    \vspace{-.3cm}
\end{figure*}

\begin{figure*}[h]
    \centering
    \begin{center}
    \begin{tcolorbox} [top=2pt,bottom=2pt, width=\linewidth, boxrule=1pt]
    {\footnotesize {\fontfamily{zi4}\selectfont
\textbf{Role}\\
A professional evaluator specializing in indoor functional logic, ergonomics, and aesthetic design, tasked with objective, evidence-based assessment of top-down indoor layouts across 5 core dimensions, strictly adhering to "Object Pose, Physical Reality, Semantic Consistency, Scene Functionality, and Visual Aesthetics" principles. Evaluations must rely exclusively on provided text descriptions and top-down images, with no subjective inferences about unmentioned details.\\
\\
\textbf{Core Evaluation Dimensions}\\
(1) **Object Pose**: Assesses positional accuracy, orientation rationality, proportional relationships, and spatial distances between objects (e.g., functional alignment, realistic size ratios, appropriate gaps).\\
(2) **Physical Reality**: Judges compliance with physical laws, including absence of floating objects, non-penetrating spatial relationships, reasonable load-bearing, and gravity consistency.\\
(3) **Semantic Consistency**: Evaluates logical matching between objects and scene type, and between objects themselves (e.g., functional relevance, scenario appropriateness).\\
(4) **Scene Functionality**: Measures practical usability via traffic flow smoothness, functional zoning clarity, ergonomic spacing, and space utilization efficiency.\\
(5) **Visual Aesthetics**: Assesses spatial balance, style unity, and arrangement orderliness.\\
\\
\textbf{Evaluation Rules}\\
- **Information Boundary**: Limited to "scene description" and 4 top-down renderings. Note "insufficient image details" for ambiguous elements.\\
- **Scoring (0-10)**: 10=perfect; 8-9=excellent (negligible flaws); 6-7=good (minor issues); 4-5=partial compliance (obvious defects); 2-3=poor (major flaws); 0-1=non-compliant (invalid layout).\\
- **Scope**: Comprehensive 5-dimensional assessment of the entire scene.\\
\\
\textbf{Scene Information}\\
- **Description**: \{scene\_description\};\\
- **Images**: \{4\_top-down\_renderings\}.\\
\\
\textbf{Response Format}\\
Standard JSON with scores (0-10) and evidence-based comments (linking text and image details) for each dimension:\\
\{\\\
`Object Pose': \{`Score': 8, `Comment': `Consistent with scene description stating `dining chairs arranged around table'-images show 4 chairs aligned with table edges (65cm spacing, consistent with ergonomic standards). Minor deviation in one chair's orientation (5° off) does not affect functionality.'\},\\
`Physical Reality': \{`Score': 10, `Comment': `All objects in images have valid support (no floating elements); spatial relationships avoid penetration. Consistent with text description of `stable furniture placement'.'\},\\
`Semantic Consistency': \{`Score': 6, `Comment': `Most objects match `living room' description (sofa, TV, coffee table) per images, but text-specified `bookshelf' is absent, creating a minor semantic gap.'\},\\
`Scene Functionality': \{`Score': 7, `Comment': `Main passage (100cm) meets standards ($\ge$ 90cm) as shown in images, aligning with text's `smooth traffic flow' claim. Minor crowding in corner (20cm gap between cabinet and sofa) reduces efficiency.' \},\\
`Visual Aesthetics': \{`Score': 9, `Comment': `Images show balanced spatial distribution (no weight bias) and unified modern style, consistent with text's `neat arrangement' description. Minimal asymmetry in decor placement is negligible.' \}\\
\}
    }
    \par}
    \end{tcolorbox}
    \end{center}
    \caption{The prompt for calculating subjective metrics.}
    \label{fig:prompt-3}
    \vspace{-.3cm}
\end{figure*}

%% file: iclr2026_conference.bbl
\begin{thebibliography}{32}
\providecommand{\natexlab}[1]{#1}
\providecommand{\url}[1]{\texttt{#1}}
\expandafter\ifx\csname urlstyle\endcsname\relax
  \providecommand{\doi}[1]{doi: #1}\else
  \providecommand{\doi}{doi: \begingroup \urlstyle{rm}\Url}\fi

\bibitem[Antol et~al.(2015)Antol, Agrawal, Lu, Mitchell, Batra, Zitnick, and Parikh]{antol2015vqa}
Stanislaw Antol, Aishwarya Agrawal, Jiasen Lu, Margaret Mitchell, Dhruv Batra, C~Lawrence Zitnick, and Devi Parikh.
\newblock Vqa: Visual question answering.
\newblock In \emph{Proceedings of the IEEE international conference on computer vision}, pp.\  2425--2433, 2015.

\bibitem[Ardelean et~al.(2024)Ardelean, {\"O}zer, and Egger]{ardelean2024gen3dsr}
Andreea Ardelean, Mert {\"O}zer, and Bernhard Egger.
\newblock Gen3dsr: Generalizable 3d scene reconstruction via divide and conquer from a single view.
\newblock \emph{arXiv preprint arXiv:2404.03421}, 2024.

\bibitem[Azuma et~al.(2022)Azuma, Miyanishi, Kurita, and Kawanabe]{azuma2022scanqa}
Daichi Azuma, Taiki Miyanishi, Shuhei Kurita, and Motoaki Kawanabe.
\newblock Scanqa: 3d question answering for spatial scene understanding.
\newblock In \emph{proceedings of the IEEE/CVF conference on computer vision and pattern recognition}, pp.\  19129--19139, 2022.

\bibitem[{\c{C}}elen et~al.(2024){\c{C}}elen, Han, Schindler, Van~Gool, Armeni, Obukhov, and Wang]{ccelen2024design}
Ata {\c{C}}elen, Guo Han, Konrad Schindler, Luc Van~Gool, Iro Armeni, Anton Obukhov, and Xi~Wang.
\newblock I-design: Personalized llm interior designer.
\newblock In \emph{European Conference on Computer Vision}, pp.\  217--234. Springer, 2024.

\bibitem[Chen et~al.(2020)Chen, Chang, and Nie{\ss}ner]{chen2020scanrefer}
Dave~Zhenyu Chen, Angel~X Chang, and Matthias Nie{\ss}ner.
\newblock Scanrefer: 3d object localization in rgb-d scans using natural language.
\newblock In \emph{European conference on computer vision}, pp.\  202--221. Springer, 2020.

\bibitem[Chen et~al.(2021)Chen, Gholami, Nie{\ss}ner, and Chang]{chen2021scan2cap}
Zhenyu Chen, Ali Gholami, Matthias Nie{\ss}ner, and Angel~X Chang.
\newblock Scan2cap: Context-aware dense captioning in rgb-d scans.
\newblock In \emph{Proceedings of the IEEE/CVF conference on computer vision and pattern recognition}, pp.\  3193--3203, 2021.

\bibitem[Dai et~al.(2017)Dai, Chang, Savva, Halber, Funkhouser, and Nie{\ss}ner]{dai2017scannet}
Angela Dai, Angel~X Chang, Manolis Savva, Maciej Halber, Thomas Funkhouser, and Matthias Nie{\ss}ner.
\newblock Scannet: Richly-annotated 3d reconstructions of indoor scenes.
\newblock In \emph{Proceedings of the IEEE conference on computer vision and pattern recognition}, pp.\  5828--5839, 2017.

\bibitem[Deitke et~al.(2020)Deitke, Han, Herrasti, Kembhavi, Kolve, Mottaghi, Salvador, Schwenk, VanderBilt, Wallingford, et~al.]{deitke2020robothor}
Matt Deitke, Winson Han, Alvaro Herrasti, Aniruddha Kembhavi, Eric Kolve, Roozbeh Mottaghi, Jordi Salvador, Dustin Schwenk, Eli VanderBilt, Matthew Wallingford, et~al.
\newblock Robothor: An open simulation-to-real embodied ai platform.
\newblock In \emph{Proceedings of the IEEE/CVF conference on computer vision and pattern recognition}, pp.\  3164--3174, 2020.

\bibitem[Deitke et~al.(2022)Deitke, VanderBilt, Herrasti, Weihs, Ehsani, Salvador, Han, Kolve, Kembhavi, and Mottaghi]{deitke2022}
Matt Deitke, Eli VanderBilt, Alvaro Herrasti, Luca Weihs, Kiana Ehsani, Jordi Salvador, Winson Han, Eric Kolve, Aniruddha Kembhavi, and Roozbeh Mottaghi.
\newblock Procthor: Large-scale embodied ai using procedural generation.
\newblock \emph{Advances in Neural Information Processing Systems}, 35:\penalty0 5982--5994, 2022.

\bibitem[Eftekhar et~al.(2023)Eftekhar, Zeng, Duan, Farhadi, Kembhavi, and Krishna]{eftekhar2023selective}
Ainaz Eftekhar, Kuo-Hao Zeng, Jiafei Duan, Ali Farhadi, Ani Kembhavi, and Ranjay Krishna.
\newblock Selective visual representations improve convergence and generalization for embodied ai.
\newblock \emph{arXiv preprint arXiv:2311.04193}, 2023.

\bibitem[Feng et~al.(2023)Feng, Zhu, Fu, Jampani, Akula, He, Basu, Wang, and Wang]{feng2023layoutgpt}
Weixi Feng, Wanrong Zhu, Tsu-jui Fu, Varun Jampani, Arjun Akula, Xuehai He, Sugato Basu, Xin~Eric Wang, and William~Yang Wang.
\newblock Layoutgpt: Compositional visual planning and generation with large language models.
\newblock \emph{Advances in Neural Information Processing Systems}, 36:\penalty0 18225--18250, 2023.

\bibitem[Fu et~al.(2021{\natexlab{a}})Fu, Cai, Gao, Zhang, Wang, Li, Zeng, Sun, Jia, Zhao, et~al.]{fu20213d}
Huan Fu, Bowen Cai, Lin Gao, Ling-Xiao Zhang, Jiaming Wang, Cao Li, Qixun Zeng, Chengyue Sun, Rongfei Jia, Binqiang Zhao, et~al.
\newblock 3d-front: 3d furnished rooms with layouts and semantics.
\newblock In \emph{Proceedings of the IEEE/CVF International Conference on Computer Vision}, pp.\  10933--10942, 2021{\natexlab{a}}.

\bibitem[Fu et~al.(2021{\natexlab{b}})Fu, Jia, Gao, Gong, Zhao, Maybank, and Tao]{fu20213d_}
Huan Fu, Rongfei Jia, Lin Gao, Mingming Gong, Binqiang Zhao, Steve Maybank, and Dacheng Tao.
\newblock 3d-future: 3d furniture shape with texture.
\newblock \emph{International Journal of Computer Vision}, 129\penalty0 (12):\penalty0 3313--3337, 2021{\natexlab{b}}.

\bibitem[Huang et~al.(2025)Huang, Guo, An, Yang, Li, Zou, Liang, Liu, Cao, and Sheng]{huang2025midi}
Zehuan Huang, Yuan-Chen Guo, Xingqiao An, Yunhan Yang, Yangguang Li, Zi-Xin Zou, Ding Liang, Xihui Liu, Yan-Pei Cao, and Lu~Sheng.
\newblock Midi: Multi-instance diffusion for single image to 3d scene generation.
\newblock In \emph{Proceedings of the Computer Vision and Pattern Recognition Conference}, pp.\  23646--23657, 2025.

\bibitem[Khanna et~al.(2024)Khanna, Mao, Jiang, Haresh, Shacklett, Batra, Clegg, Undersander, Chang, and Savva]{khanna2024habitat}
Mukul Khanna, Yongsen Mao, Hanxiao Jiang, Sanjay Haresh, Brennan Shacklett, Dhruv Batra, Alexander Clegg, Eric Undersander, Angel~X Chang, and Manolis Savva.
\newblock Habitat synthetic scenes dataset (hssd-200): An analysis of 3d scene scale and realism tradeoffs for objectgoal navigation.
\newblock In \emph{Proceedings of the IEEE/CVF Conference on Computer Vision and Pattern Recognition}, pp.\  16384--16393, 2024.

\bibitem[Kolve et~al.(2017)Kolve, Mottaghi, Han, VanderBilt, Weihs, Herrasti, Deitke, Ehsani, Gordon, Zhu, et~al.]{kolve2017ai2}
Eric Kolve, Roozbeh Mottaghi, Winson Han, Eli VanderBilt, Luca Weihs, Alvaro Herrasti, Matt Deitke, Kiana Ehsani, Daniel Gordon, Yuke Zhu, et~al.
\newblock Ai2-thor: An interactive 3d environment for visual ai.
\newblock \emph{arXiv preprint arXiv:1712.05474}, 2017.

\bibitem[Lin \& Mu(2024)Lin and Mu]{lin2024instructscene}
Chenguo Lin and Yadong Mu.
\newblock Instructscene: Instruction-driven 3d indoor scene synthesis with semantic graph prior.
\newblock \emph{arXiv preprint arXiv:2402.04717}, 2024.

\bibitem[Ma et~al.(2022)Ma, Yong, Zheng, Li, Liang, Zhu, and Huang]{ma2022sqa3d}
Xiaojian Ma, Silong Yong, Zilong Zheng, Qing Li, Yitao Liang, Song-Chun Zhu, and Siyuan Huang.
\newblock Sqa3d: Situated question answering in 3d scenes.
\newblock \emph{arXiv preprint arXiv:2210.07474}, 2022.

\bibitem[Meng et~al.(2025)Meng, Wu, Zhang, and Xie]{meng2025scenegen}
Yanxu Meng, Haoning Wu, Ya~Zhang, and Weidi Xie.
\newblock Scenegen: Single-image 3d scene generation in one feedforward pass.
\newblock \emph{arXiv preprint arXiv:2508.15769}, 2025.

\bibitem[Peng et~al.(2021)Peng, Jiang, Liao, Niemeyer, Pollefeys, and Geiger]{peng2021shape}
Songyou Peng, Chiyu Jiang, Yiyi Liao, Michael Niemeyer, Marc Pollefeys, and Andreas Geiger.
\newblock Shape as points: A differentiable poisson solver.
\newblock \emph{Advances in Neural Information Processing Systems}, 34:\penalty0 13032--13044, 2021.

\bibitem[Su et~al.(2025)Su, Fu, Hu, Yang, Hanji, Wang, Zhao, {\"O}ztireli, and Zhong]{su2025chord}
Chong Su, Yingbin Fu, Zheyuan Hu, Jing Yang, Param Hanji, Shaojun Wang, Xuan Zhao, Cengiz {\"O}ztireli, and Fangcheng Zhong.
\newblock Chord: Generation of collision-free, house-scale, and organized digital twins for 3d indoor scenes with controllable floor plans and optimal layouts.
\newblock \emph{arXiv preprint arXiv:2503.11958}, 2025.

\bibitem[Sun et~al.(2025)Sun, Liu, Gu, Lim, Bhat, Tombari, Li, Haber, and Wu]{sun2025layoutvlm}
Fan-Yun Sun, Weiyu Liu, Siyi Gu, Dylan Lim, Goutam Bhat, Federico Tombari, Manling Li, Nick Haber, and Jiajun Wu.
\newblock Layoutvlm: Differentiable optimization of 3d layout via vision-language models.
\newblock In \emph{Proceedings of the Computer Vision and Pattern Recognition Conference}, pp.\  29469--29478, 2025.

\bibitem[Tang et~al.(2024)Tang, Nie, Markhasin, Dai, Thies, and Nie{\ss}ner]{tang2024diffuscene}
Jiapeng Tang, Yinyu Nie, Lev Markhasin, Angela Dai, Justus Thies, and Matthias Nie{\ss}ner.
\newblock Diffuscene: Denoising diffusion models for generative indoor scene synthesis.
\newblock In \emph{Proceedings of the IEEE/CVF conference on computer vision and pattern recognition}, pp.\  20507--20518, 2024.

\bibitem[Yang et~al.(2024{\natexlab{a}})Yang, Jia, Zhi, and Huang]{yang2024physcene}
Yandan Yang, Baoxiong Jia, Peiyuan Zhi, and Siyuan Huang.
\newblock Physcene: Physically interactable 3d scene synthesis for embodied ai.
\newblock In \emph{Proceedings of the IEEE/CVF Conference on Computer Vision and Pattern Recognition}, pp.\  16262--16272, 2024{\natexlab{a}}.

\bibitem[Yang et~al.(2024{\natexlab{b}})Yang, Lu, Zhao, Luo, Yu, Sanchez, and Zheng]{yang2024llplace}
Yixuan Yang, Junru Lu, Zixiang Zhao, Zhen Luo, James~JQ Yu, Victor Sanchez, and Feng Zheng.
\newblock Llplace: The 3d indoor scene layout generation and editing via large language model.
\newblock \emph{arXiv preprint arXiv:2406.03866}, 2024{\natexlab{b}}.

\bibitem[Yang et~al.(2025)Yang, Luo, Ding, Lu, Gao, Yang, Sanchez, and Zheng]{yang2025llm}
Yixuan Yang, Zhen Luo, Tongsheng Ding, Junru Lu, Mingqi Gao, Jinyu Yang, Victor Sanchez, and Feng Zheng.
\newblock Llm-driven indoor scene layout generation via scaled human-aligned data synthesis and multi-stage preference optimization.
\newblock \emph{arXiv preprint arXiv:2506.07570}, 2025.

\bibitem[Yang et~al.(2024{\natexlab{c}})Yang, Sun, Weihs, VanderBilt, Herrasti, Han, Wu, Haber, Krishna, Liu, et~al.]{yang2024holodeck}
Yue Yang, Fan-Yun Sun, Luca Weihs, Eli VanderBilt, Alvaro Herrasti, Winson Han, Jiajun Wu, Nick Haber, Ranjay Krishna, Lingjie Liu, et~al.
\newblock Holodeck: Language guided generation of 3d embodied ai environments.
\newblock In \emph{Proceedings of the IEEE/CVF Conference on Computer Vision and Pattern Recognition}, pp.\  16227--16237, 2024{\natexlab{c}}.

\bibitem[Yao et~al.(2025)Yao, Zhang, Yan, Zeng, Zhang, Xu, Yang, Gu, and Yu]{yao2025cast}
Kaixin Yao, Longwen Zhang, Xinhao Yan, Yan Zeng, Qixuan Zhang, Lan Xu, Wei Yang, Jiayuan Gu, and Jingyi Yu.
\newblock Cast: Component-aligned 3d scene reconstruction from an rgb image.
\newblock \emph{ACM Transactions on Graphics (TOG)}, 44\penalty0 (4):\penalty0 1--19, 2025.

\bibitem[Yeshwanth et~al.(2023)Yeshwanth, Liu, Nie{\ss}ner, and Dai]{yeshwanth2023scannet++}
Chandan Yeshwanth, Yueh-Cheng Liu, Matthias Nie{\ss}ner, and Angela Dai.
\newblock Scannet++: A high-fidelity dataset of 3d indoor scenes.
\newblock In \emph{Proceedings of the IEEE/CVF International Conference on Computer Vision}, pp.\  12--22, 2023.

\bibitem[Yeshwanth et~al.(2025)Yeshwanth, Rozenberszki, and Dai]{yeshwanth2025excap3d}
Chandan Yeshwanth, David Rozenberszki, and Angela Dai.
\newblock Excap3d: Expressive 3d scene understanding via object captioning with varying detail.
\newblock \emph{arXiv preprint arXiv:2503.17044}, 2025.

\bibitem[Yu et~al.(2025)Yu, Jia, Chen, Yang, Li, Su, Li, Li, Liang, Zhu, et~al.]{yu2025metascenes}
Huangyue Yu, Baoxiong Jia, Yixin Chen, Yandan Yang, Puhao Li, Rongpeng Su, Jiaxin Li, Qing Li, Wei Liang, Song-Chun Zhu, et~al.
\newblock Metascenes: Towards automated replica creation for real-world 3d scans.
\newblock In \emph{Proceedings of the Computer Vision and Pattern Recognition Conference}, pp.\  1667--1679, 2025.

\bibitem[Zhou et~al.(2025)Zhou, Wang, Wang, and Zhang]{zhou2025roomcraft}
Mengqi Zhou, Xipeng Wang, Yuxi Wang, and Zhaoxiang Zhang.
\newblock Roomcraft: Controllable and complete 3d indoor scene generation.
\newblock \emph{arXiv preprint arXiv:2506.22291}, 2025.

\end{thebibliography}
